\documentclass[lettersize,journal]{IEEEtran}

\pdfoutput=1

\usepackage{amsmath,amsfonts}
\usepackage{algorithmic}
\usepackage{algorithm}
\usepackage{array}
\usepackage[caption=false,font=normalsize,labelfont=sf,textfont=sf]{subfig}
\usepackage{textcomp}
\usepackage{stfloats}
\usepackage{url}
\usepackage{verbatim}
\usepackage{graphicx}
\usepackage{cite}
\usepackage{textcomp}
\usepackage{tablefootnote}
\usepackage{physics}
\usepackage[edges]{forest}
\usepackage{tabularx, booktabs, array}
\tikzset{%
    parent/.style =          {align=center,text width=5cm,rounded corners=3pt},
    child/.style =           {align=center,text width=2.5cm,rounded corners=3pt},
    grandchild/.style =      {align=center,text width=2cm,rounded corners=3pt},
    greatgrandchild/.style = {align=center,text width=3.5cm,rounded corners=3pt},
    referenceblock/.style =  {align=left,text width=8cm,rounded corners=2pt}
}

\usepackage{bbm}
\usepackage{amsthm}
\usepackage{hyperref}
\usepackage{makecell}
\hypersetup{hidelinks, colorlinks=true,allcolors=black,pdfstartview=Fit,breaklinks=true}
\usepackage{longtable}
\usepackage{multirow}
\usepackage{amssymb}
\usepackage{booktabs}

\begin{document}

\title{Uncertainty Quantification on Graph Learning: A Survey}
\author{
Chao Chen$^{\dagger}$, Chenghua Guo$^{\dagger}$, Rui Xu$^{\dagger}$, Jiujiu Chen,\thanks{${\dagger}$ Three authors are listed in alphabetical order with equal contributions to this work. 
Specifically, Chao primarily focused on Sec. \ref{sec:preliminaries}, Sec. \ref{sec:outlier_representation}, Sec. \ref{sec:ood_handling}, and Sec. \ref{sec:conformal_prediction}; 
Chenghua primarily worked on Sec. \ref{sec:method_uncertainty_representation} and \ref{subsec: bayesian_inference_techniques}; 
and Rui primarily handled Sec. \ref{sec:source_of_uncertainty}, Sec. \ref{sec:calibration}, Sec. \ref{sec:conformal_prediction}, and Sec. \ref{sec:uq_evaluation}. Jiujiu primarily focused on all subsections of GFMs.}
Xiangwen Liao, Xi Zhang,~\IEEEmembership{Member,~IEEE}, \\
Sihong Xie$^{\ast}$,~\IEEEmembership{Senior member,~IEEE} \thanks{${\ast}$Corresponding author: sihongxie@hkust-gz.edu.cn}, 
Hui Xiong,~\IEEEmembership{Fellow,~IEEE}, 
Philip S. Yu~\IEEEmembership{Fellow,~IEEE}
}

\maketitle

\begin{abstract}
Graphical models
have demonstrated their exceptional capabilities across numerous applications. 
However, their performance, confidence, and trustworthiness are often limited by the inherent randomness in data generation and the lack of knowledge to accurately model real-world complexities. 
There has been increased interest in developing uncertainty quantification (UQ) techniques tailored to graphical models. 
In this survey, we systematically examine existing works on UQ for graphical models. 
This survey distinguishes itself from most existing UQ surveys by specifically concentrating on graphical models, including 
graph neural networks and graph foundation models. 
We organize the literature along two complementary dimensions:
uncertainty representation and uncertainty handling. 
By synthesizing both established methodologies and emerging trends, we aim to bridge gaps in understanding key challenges and opportunities in UQ for graphical models, inspiring researchers on graphical models or uncertainty quantification to make further advancements at the cross of the two fields.
\end{abstract}

\begin{IEEEkeywords}
Surveys, Uncertainty, Graph neural networks
\end{IEEEkeywords}
\maketitle

\section{Introduction}
Graphical models have emerged as fundamental tools in machine learning due to their capability to capture and represent complex relational dynamics among variables. 
These models have found extensive applications, such as
spam detection in online review systems \cite{rayana2015collective,noekhah2020opinion}, 
relationship extraction in social networks 
\cite{agarwal2009social,kipf2016semi}, 
user interest exploration in recommendation systems 
\cite{gupta2020calibration,guo2017calibration,huang2022mbct}, 
and drug discovery in protein-protein interaction networks 
\cite{reau2023deeprank,jha2022prediction}.
Rather than assuming variables to be independently and identically distributed (i.i.d.), graphical models reveal intricate relationships and conditional dependencies. 
By considering both individual variables and their interconnections, graphical models improve predictive performance in complex data environments and tasks. 

Graphical models are used for a broad range of downstream tasks, especially node- and graph-level classification, and link prediction. 
However, uncertainties are ubiquitous and can arise from various sources, including inherent data randomness, learning algorithm imperfections, and unexpected test data distributions \cite{wang2024uncertainty}.
These uncertainties hinder the deployment of graphical models in high-stakes domains where precise risk assessment and decision-making under uncertainty are vital,
including healthcare \cite{seoni2023application,jackson2010structural}, autonomous driving \cite{tang2023uncertainty,suk2025uncertainty}, and natural language generation~\cite{ma2025estimatingllmuncertaintyevidence}.

Uncertainty quantification (UQ) aims to associate predictions with calibrated measures of confidence by distinct ways, such as identifying out-of-distribution data, supporting safety-critical decision-making via risk-sensitive objectives, or improving model selection and debugging through uncertainty-aware learning. 
Motivated by these needs, a growing body of work has developed UQ techniques for graphical models.

This survey explores uncertainties in 
graph neural networks (GNNs) and graph foundation models (GFMs). 
GNNs \cite{kipf2016semi,reau2023deeprank} extend the successes of deep learning to graph-structured data but face uncertainty challenges such as noisy links, missing nodes, or mislabeled data. 
Recent developments have incorporated UQ into the architectures and/or training algorithms of GNNs, improving their robustness and interpretability. 
Inspired by the pre-training strategies in large language models (LLMs), 
GFMs are proposed to extend graph learning to the paradigm of graph pre-training and task-specific adaptation \cite{liu2025graph}, but the research on UQ in GFMs remains scarce. 

\begin{figure*}
    \centering
    \includegraphics[width=\textwidth]{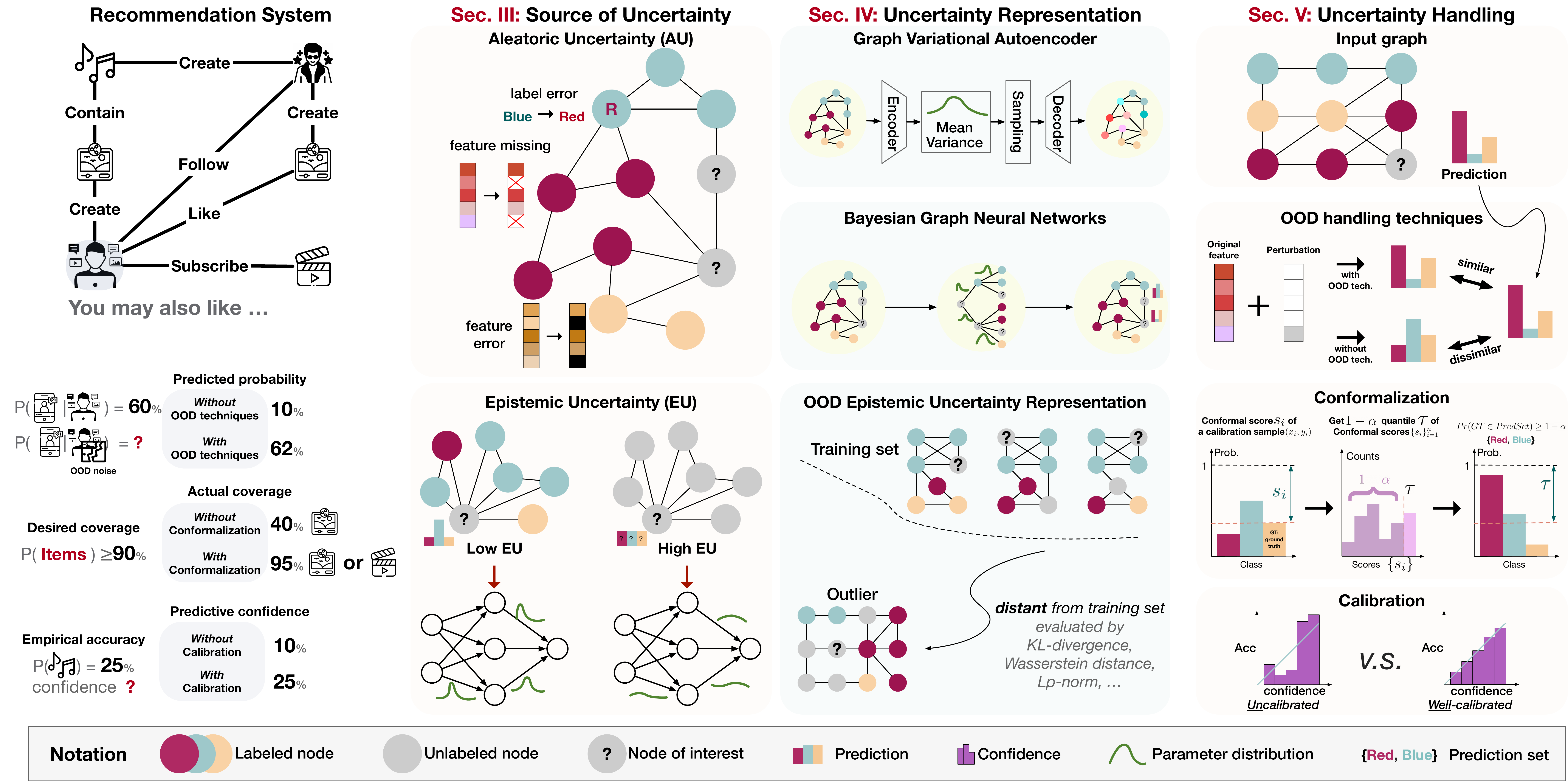}
    \caption{An example from a recommendation system, where the user of interest (in gray) links to other users and contents with various relationships. 
    A model's predictions for the user's preference under uncertainty are shown below, with or without uncertainty handling methods.
    Then, as detailed in Sec. \ref{sec:source_of_uncertainty}, the \textit{source} of uncertainty can be decomposed into aleatoric and epistemic uncertainties. 
    Last, we demonstrate representative methods for uncertainty \textit{representation} (in Sec. \ref{sec:method_uncertainty_representation}) and \textit{handling} (in Sec. \ref{sec:method_uncertainty_handling}), respectively.
    }
    \label{fig:uq_intro}
\end{figure*}

Existing surveys on UQ cover broad overviews of methods \cite{abdar2021review,gawlikowski2023survey,mena2021survey,wang2020survey} and focused studies on particular areas. 
These areas include scientific machine learning \cite{psaros2023uncertainty}, sources of uncertainty \cite{gruber2023sources,hullermeier2021aleatoric}, optimization under uncertainty \cite{ning2019optimization}, and noise in machine learning \cite{gupta2019dealing}.
%
Other surveys examine specific paradigms, such as integrating differential equations with GNNs, but discuss uncertainty in that context only briefly \cite{liu2025graph}.
However, comprehensive and systematic surveys specifically dedicated to UQ for graph learning remain scarce. 

A survey closely related to ours is that of Wang et al. \cite{wang2024uncertainty}, which provides a broad overview of UQ in graph learning. 
However, they primarily focus on downstream tasks and applications related to GNN uncertainty, rather than on the theoretical foundations of UQ methods. 
In contrast, our work places greater emphasis on the mathematical underpinnings of each UQ approach, presenting key formulas and core concepts that offer a principled perspective of uncertainty in graphs. 
Moreover, whereas they categorize methods by ``single models versus ensemble models'' and ``deterministic models versus models with random parameters,'' which are model‑centric, we instead organize UQ methods into \textbf{``uncertainty representation''} (how uncertainty is modeled) and \textbf{``uncertainty handling''} (how uncertainty is utilized or mitigated), which more clearly reflects the roles of different techniques. 
Finally, our survey incorporates advanced UQ for graph learning, such as frameworks based on stochastic differential equations, which are not covered in Wang et al. \cite{wang2024uncertainty}.
Fig. \ref{fig:uq_intro} shows an example of uncertainty on a recommendation system, 
and highlights three key components of the survey: source, representation, and handling of uncertainties.
We structure this survey as follows,
we first provide the preliminary knowledge concerning
graphs in Sec. \ref{sec:preliminaries}, 
and introduce the sources of uncertainty in Sec. \ref{sec:source_of_uncertainty}. 
We then discuss the existing studies for uncertainty representation and uncertainty handling in Sec. \ref{sec:method_uncertainty_representation} and Sec. \ref{sec:method_uncertainty_handling}, respectively. 
Furthermore, we demonstrate uncertainty evaluation protocols in Sec. \ref{sec:uq_evaluation}.
Finally, future directions and the conclusion of the survey are included in Sec. \ref{sec:future_direction} and Sec. \ref{sec:conclusion}, respectively.

\section{Preliminaries}
\label{sec:preliminaries}


\subsection{Graph Neural Networks}
Consider a graph $G=(\mathcal{V},\mathcal{E})$ consisting of $n$ nodes $\mathcal{V}=\{ V_1, \dots, V_n \}$, 
and each $V_i \in \mathcal{V}$ is associated with a feature vector $X_i$, where the feature matrix $\mathbf{X}$ collects all node features.
Each edge in the edge set $\mathcal{E}$ is a relationship between two nodes. 
The set $\mathcal{N}(V_i) = \{ V_j: (V_i, V_j)\in\mathcal{E} \}$ collects direct neighbors of $V_i$. 
An adjacency matrix $\mathbf{A}$ represents the connections among nodes. 
Graph Neural Network (GNN) is a deep learning technique designed to  
learn complex relationships in graphs.
Formally, 
the intermediate representation $\mathbf{H}^{(l)}$ of the $l$-th layer from an $L$-layers GNN is updated by
\begin{equation}
    \mathbf{H}^{(l)}=\sigma\left(\tilde{\mathbf{A}}\mathbf{H}^{(l-1)}\theta^{(l-1)}\right),
\end{equation}
where $\sigma(\cdot)$ is an activation function.
$\mathbf{H}^{(0)}=\mathbf{X}$ is the input feature matrix,
and $\mathbf{H}^{(L)}$ from the last layer could be used for downstream prediction tasks.
$\theta^{(l)}$ is model parameter of layer $l$.
$\tilde{\mathbf{A}}=\mathbf{D}^{-1/2}\mathbf{A}_{+I}\mathbf{D}^{-1/2}$ is a transformation of $\mathbf{A}$ \cite{kipf2016semi}, 
where $\mathbf{A}_{+I}=\mathbf{A}+\mathbf{I}$ and $\mathbf{D}$ is the degree matrix of $\mathbf{A}_{+I}$.

\subsection{Graph Foundation Models}
The remarkable success of foundation models in language and vision has motivated their extension to graph machine learning \cite{wang2023visionllm, mao2024position}. Therefore, graph foundation models (GFMs) are designed to achieve broad adaptability across diverse domains and tasks, and also expected to exhibit emergence and homogenization \cite{liu2025graph}. Recent studies generally classify GFMs into three paradigms: GNN-based \cite{cui2024prompt}, LLM-based \cite{zhu2025llm, ye2024language}, and GNN+LLM-based GFMs \cite{zhu2025graphclip}. GNN-based approaches employ GNNs as backbone models pretrained on various graphs, but suffer from cross-domain feature alignment issues during downstream adaptation. LLM-based methods transform graph into textual representations, leveraging natural language to unify domains and tasks. Although LLMs exhibit strong generalization in graph tasks, they are inherently limited in capturing complex graph structures~\cite{chen2024text, zhou2025each}. To alleviate the limitation, GNN+LLM-based GFMs integrate the structural reasoning capability of GNNs with the semantic generalization power of LLMs, thereby combining the strengths of both paradigms. 
\section{Sources of Uncertainty}
\label{sec:source_of_uncertainty}
The total uncertainty in machine learning can be decomposed into \textbf{aleatoric uncertainty} (AU) and \textbf{epistemic uncertainty} (EU) based on their sources~\cite{der2009aleatory}\footnote{Recent studies have challenged the traditional binary classification of uncertainty into aleatoric and epistemic types, suggesting that this dichotomy may be overly simplistic~\cite{kaiser2025uncertainsam, kirchhof2025position}.}.
Aleatoric uncertainty, which is irreducible with increasing amount of data, represents the randomness inherent to the data generation process~\cite{hacking2006emergence}. Epistemic uncertainty can be reduced by obtaining more knowledge or data~\cite{kendall2017uncertainties, hullermeier2021aleatoric}. Valdenegro-Toro et al.~\cite{valdenegrotoro2022deeper} introduce techniques for the disentanglement of AU and EU, and Munikoti et al.~\cite{munikoti2023general} formulate mathematical expressions of these two kinds of uncertainty in the context of GNNs.



We mathematically define the two sources of uncertainty and how they impact prediction reliability. Let $X\in\mathcal{X}\subseteq\mathbb{R}^d$ and $Y\in\mathcal{Y}\subseteq\mathbb{R}$ denote the input and output variables with realizations $x$ and $y$, respectively. The underlying data distribution is $P(X, Y)$ and the data generation mechanism is $f_0$ such that $y=f_0(x)$. Therefore, the joint distribution can be written as $P(X,Y)= P(X,f_0(X))$.

\subsection{Aleatoric Uncertainty}
When the ground truth mapping function \( f_0 \) is conditioned on \( X = x \), aleatoric uncertainty (AU) can be quantified through the variance of the conditional label distribution \( P(Y \mid x) \). The dispersion of this distribution quantifies the AU and is often measured by the conditional variance of \( P(Y \mid x)\).
Gruber et al.~\cite{gruber2023sources} identify four sources of AU as follows.

\textbf{Omitted Features} are variables excluded during model training and prediction. These omitted features contribute to AU, since collecting more data without these missing features does not reduce the uncertainty they introduce. In graph-structured data, omitted features may correspond to unobserved node attributes, missing edges, or partially observed subgraph structures. The omission is investigated in the context of causal models~\cite{chernozhukov2021omitted}, confounder analysis~\cite{busenbark2022omitted}, and model sensitivity analysis~\cite{cinelli2020making}. The omission may result from variables being unobservable or overlooked in data collection. 


\textbf{Feature Errors} represent inaccurate feature values in collected data, which are likely to occur with improper measurements~\cite{gupta2019dealing}. For example, in traffic graph data, edge attributes such as speed or traffic volume may be inaccurately estimated due to low-frequency sampling, sensor noise, or malfunctioning devices. These errors can cause biases and increase variance in observed features. 

\textbf{Label Errors} mean the labels can be incorrect. These errors can be introduced by human labeling with subjectivity. For instance, in node classification tasks on social networks, users may be incorrectly labeled as malicious actors. Label errors in training data can mislead gradient descent by causing incorrect update directions, and those in test sets will improperly evaluate models' performance. 
In this case, uncertainty in \( Y \) arises not only from inherent ambiguity in the input-output relationship, but also from label noise. This contributes to AU, since it cannot be reduced merely by increasing the number of samples without improving label quality.

\textbf{Missing Data} indicates some entries in the dataset are incomplete~\cite{little2019statistical}. This differs from omitted features, which refer to entire variables that are excluded from the dataset, rather than individual missing values within an included feature. Let $M \in \{0,1\}$ denote the missingness indicator, where $M=0$ corresponds to complete samples and $M=1$ corresponds to samples with missing values.
The influence from incompleteness can be expressed by
\begin{equation}
    P(Y|X,M=0)=\frac{P(M=0|Y,X)}{P(M=0|X)}P(Y|X).
\end{equation}
This equation reveals that missing data can introduce selection bias into the conditional distribution of complete samples \( P({Y\mid X, M=0} )\), which deviates from the true distribution \( P({Y \mid X}) \). The uncertainty arising from this bias is aleatoric in nature, as it stems from incomplete observations that cannot be resolved simply by collecting more data unless the missingness mechanism is addressed or corrected.

\subsection{Epistemic Uncertainty}
Epistemic uncertainty (EU) is from the lack of knowledge and can be categorized into model, approximation, and Out-of-Distribution (OOD) uncertainties~\cite{hullermeier2021aleatoric}. 

\textbf{Model Uncertainty} refers to the difference between the ground truth mapping $f_0 \in \mathcal{F}_0$ and the best possible predictor $f \in \mathcal{F}$. 
It is impossible to exhaustively explore the entire function space $\mathcal{F}_0$. Consequently, the learned function $f\in\mathcal{F}\subset\mathcal{F}_0$, which induces a gap between $f$ and the target function $f_0$. 
Thereby, we can represent model uncertainty by
$\text{dist}( P(X,Y), P(X,f(X)))$
where $\text{dist}(\cdot,\cdot)$ is a user-specified discrepancy measure.

\textbf{Approximation Uncertainty} is defined as the uncertainty due to the limited number of $N$ training samples. Since the population distribution $P({X,Y})$ is not accessible in practice, the induced predictor trained on finite samples is defined as 
\begin{equation}
    \hat{f}=\arg\min_{f\in\mathcal{F}}\frac{1}{N}\sum_{i=1}^{N}l(y_i,f(x_i)),
\end{equation}
which allows us to quantify approximation uncertainty as $\text{dist}(P({X,Y}), P(X,\hat{f}(X)) )$.

\textbf{OOD Uncertainty} arises at the inference stage when the deployed model encounters test data drawn from a distribution different from the training distribution~\cite{9915595,everett2022improvingoutofdistributiondetectionepistemic}. Formally, let $P({X,Y})$ and $Q({X,Y})$ denote the training and test distributions, respectively. OOD uncertainty arises as a result of a distribution shift $P({X,Y}) \neq Q({X,Y})$.
%
Since the model ${f}$ is trained to minimize the expected loss under $P({X,Y})$, its performance and confidence estimates may be unreliable when evaluated under $Q({X,Y})$. Therefore, OOD uncertainty can be formally quantified by $\text{dist}(Q({X,Y}),Q(X,f(X)) )$.
 
Model and approximation uncertainties are reducible by expanding the search space in the function space $\mathcal{F}$ and increasing the sample size $N$~\cite{lahlou2021deup,nguyen2019epistemic,huang2021quantifying}. In contrast, OOD uncertainty arises from distributional shifts between training and test data and cannot be mitigated by more data or a larger model alone. Instead, it requires robust training strategies, which will be introduced in Sec. \ref{sec:method_uncertainty_handling}, to improve generalization under shifted conditions.

\section{Methods for Uncertainty Representation}

\begin{table*}[!tb]
\centering
\caption[Uncertainty Representation in Graph Neural Networks.]
{Summary of Uncertainty Representation studies on graphs\footnotemark.}
\def\arraystretch{1.2}

\newcolumntype{Y}{>{\centering\arraybackslash}X}
\begin{tabularx}{\textwidth}{Y|Y|Y|Y|Y}
\toprule
\textbf{Type} & \textbf{Methods} & 
\textbf{Task} & \textbf{Input Space} &
\textbf{Model Space} \\

\midrule
\multirow{1}{*}{\makecell{Dirichlet}}
& GPN \cite{stadler2021graph} & Node Cls & Node feat. & - \\

\midrule
\multirow{4}{*}{Bayesian}
& VGAE \cite{kipf2016variational} & Node RL & - & Latent representation \\
& BGCNN \cite{zhang2019bayesian} & Node Cls & Graph struc. & Model parameters \\
& BUP \cite{xu2022uncertainty} & Node Cls & - & GNN messages \\
& GKDE \cite{zhao2020uncertainty} & Node Cls & - & Model parameters \\

\midrule

\multirow{1}{*}{\makecell{Gaussian Process}}
& GGP \cite{ng2018bayesian} & Node Cls & - & Latent representation \\

\midrule
\multirow{5}{*}{\makecell{OOD\\Epistemic}}
& ADNCE \cite{wu2024understanding} & Graph Cls & Node feat. \& Graph struc. & - \\
& WDRO \cite{zhang2021robust} & Clustering & Graph struc. & - \\
& DWNS \cite{dai2019adversarial} & Node Cls & Node feat. \& Graph struc. & - \\
& GCC \cite{qiu2020gcc} & Node Cls \& Graph Cls & Graph struc. & - \\

\midrule
\multirow{3}{*}{\makecell{Stochastic\\Differential \\Equations}}
& GNSD \cite{lin2024graph} & Node Cls & - & Latent representation \\
& GraphGDP \cite{huang2022graphgdp} & Graph Generation & Graph struc. & Latent representation \\
& AGGDN \cite{xing2023aggdn} & Spatiotemporal pred. & - & Latent representation \\

\bottomrule
\end{tabularx}
\label{tab:uncertainty}
\end{table*}

\label{sec:method_uncertainty_representation}
Understanding how uncertainty can be represented is fundamental for designing trustworthy graph learning algorithms.
As shown in Table \ref{tab:uncertainty}, we delve into five distinct categories of methods for uncertainty representation.
\subsection{Dirichlet-Multinomial Representation} 
\label{sec:dirichlet-motinomial}
\subsubsection{Dirichlet-Multinomial Representation Overview} \hfill\\
\indent The Dirichlet-Multinomial framework combines a Diric
hlet prior $\text{Dir}(\alpha)$ over class probabilities $\Pi$ with a Multinomial likelihood $\text{Mult}(c|\Pi)$ over observed class counts $c$. Due to conjugacy, this yields a closed-form Dirichlet posterior $\text{Dir}(\alpha + c)$, which directly induces a posterior predictive distribution. 
Predictive uncertainty can then be quantified by the posterior predictive variance or entropy computed from this distribution, without requiring parameter learning.

\footnotetext{RL is short for representation learning.}

\subsubsection{Dirichlet-Multinomial Representation for Graphs} \hfill\\
\indent SocNL \cite{yamaguchi2015socnl} is one of the early works based on Dirichlet-Multinomial distribution for graph. 
It treats the labels of a node’s neighbors as Multinomial observations and adds these label counts directly to the Dirichlet prior of the node to form the posterior.
However, this approach relies on a label-sharing assumption (neighbor labels are informative and consistent) and the mere propagation of labels (i.e., directly using the neighbors’ labels as observed data). 
NETCONF \cite{eswaran2017power} introduces the propagation of multinomial messages to support both homophily and heterophily network effects. It captures these effects by employing compatibility matrices to represent the strength of connections between nodes and then using the resulting class counts as evidence in the closed-form update.
To handle high-dimensional node features, GPN \cite{stadler2021graph} first uses an multilayer perceptron (MLP) to encode each node’s features. Then, normalizing flows \cite{rezende2015variational} are applied to compute pseudo-counts for each class as observed data. Finally, Personalized PageRank \cite{page1999pagerank} propagates these pseudo-counts to update the Dirichlet prior parameters for each node.

\underline{\textit{Summary}}.
Dirichlet–Multinomial methods leverage conjugate priors to obtain closed-form posterior updates, enabling immediate UQ by computing posterior predictive uncertainty (e.g., entropy or variance) without iterative optimization. 
The primary distinction among these methods lies in how they construct neighbor evidence, ranging from direct label propagation to learned pseudo counts from neural encodings. 
However, the conjugacy requirement constrains posteriors to remain within the Dirichlet family, 
which can be limiting for more complex scenarios such as multi-modal class uncertainty, correlated classes, or non-conjugate evidence from rich node features.

\subsection{Bayesian Representation Learning}
\label{subsec: bayesian representation learning}
\subsubsection{Bayesian Representation Learning Overview} \hfill \\
\indent Bayesian representation learning quantifies uncertainty in model parameters and learned representations.
By treating representations and parameters as probability distributions rather than deterministic point estimates, this paradigm naturally bridges representation learning with uncertainty quantification.
We focus on
variational autoencoders (VAEs) \cite{kingma2013auto} and Bayesian neural networks (BNNs) ~\cite{lampinen2001bayesian,titterington2004bayesian}. 

\textbf{VAE.}
VAEs introduce uncertainty representation by modeling data generation with fully probabilistic latent variables rather than a single deterministic code. Concretely, they assume a prior distribution $P(Z)$ over the latent variable $Z$,
typically a normal distribution, and define a likelihood
$P_{\theta}(X \mid Z)$ parameterized by a decoder $\theta$,
e.g., a Gaussian whose learned variance captures the AU inherent in the input $x$.
To infer the latent representation for input \(x\), an encoder parameterized by $\phi$ approximates the true posterior with a variational distribution $Q_{\phi}(Z \mid x)$,
typically a Gaussian whose mean and variance are parameterized depending on \(x\). Sampling \(z\) from \(Q_{\phi}(Z\mid x)\) injects stochasticity into the encoding, 
quantifying the uncertainty of the latent representation for the given observation $x$.


\textbf{BNN.}
Unlike deterministic neural networks that provide point estimates, BNNs assign distributions to the parameters $\theta$ to capture model uncertainty. 
During training, BNNs update the prior distribution $P(\theta)$ with the likelihood $P(\mathcal{D}|\theta)$ of the training set $\mathcal{D}$, and produce a posterior distribution that combines prior knowledge with observed data:
\begin{equation}
    \label{eq:Bayes_Theorem}
    P(\theta|\mathcal{D})=\frac{P(\mathcal{D}|\theta)P(\theta)}{P(\mathcal{D})}.
\end{equation}
For predictions, BNNs integrate over the posterior distribution of the parameters to account for uncertainty.
For a given test sample $x^\ast$, the predictive distribution is given by:
\begin{equation}
    \label{eq:Bayes_prediction}
    P(y^\ast | x^\ast, \mathcal{D}) = \int P(y^\ast | x^\ast, \theta) P(\theta | \mathcal{D}) d\theta.
\end{equation}

\subsubsection{Bayesian Representation Learning for Graphs} \hfill \\
\indent While GNNs lay the groundwork by leveraging local neighborhood information, their deterministic nature often falls short in capturing the inherent uncertainty of real-world graphs, such as irregular structures and noisy features. 
Bayesian graph representation learning extends the probabilistic frameworks of VAEs and BNNs to graph-structured data, capturing uncertainty arising from noisy features and incomplete structures
to produce more 
robust representations and improve downstream prediction accuracy.
Two prominent approaches have been developed: graph variational autoencoders (GVAEs) \cite{kipf2016variational,hasanzadeh2019semi,li2020dirichlet,ahn2021variational,grover2019graphite,simonovsky2018graphvae,sun2021hyperbolic} and Bayesian graph neural networks (BGNNs) \cite{zhang2019bayesian,pal2019bayesian,pal2020non,xu2022uncertainty,zhao2020uncertainty,elinas2020variational,munikoti2023general,fuchsgruber2024uncertainty}.


\textbf{GVAEs.}
VGAE \cite{kipf2016variational} is one of the earliest works to apply VAEs to graph data. In VGAE, a GCN-based encoder processes the input graph and maps each node \(V_i\) to a latent variable \(Z_i\). Instead of producing a fixed embedding, the encoder parameterized by $\phi$ defines a probabilistic representation by estimating a Gaussian distribution $\mathcal{N}$ for each node:
\begin{equation}
\label{eq:VGAE}
q_{\phi}(Z_i \mid \mathbf{X},\mathbf{A}) = \mathcal{N}\!\bigl(Z_i \mid \mu_i,\, \mathrm{diag}(\sigma_i^2)\bigr),    \end{equation}
where the mean vector \(\mu_i\) represents the central embedding and the variance \(\sigma_i^2\) quantifies the uncertainty associated with node \(V_i\)'s representation. During decoding, an inner product decoder is typically employed to reconstruct the graph structure and node features from these uncertainty-aware latent representations.

To address VGAE's limitations in capturing complex dependencies, interpretability, and isolated node handling, various extensions have been proposed: hierarchical frameworks for dependency modeling \cite{hasanzadeh2019semi}, interpretable cluster-based latent factors \cite{li2020dirichlet}, normalized embeddings for isolated nodes \cite{ahn2021variational}, iterative decoders for scalable reconstruction \cite{grover2019graphite}, and graph-level representations \cite{simonovsky2018graphvae}.
For dynamic graphs, 
VGAE is extended to model temporal dynamics 
through hierarchical variational frameworks \cite{hajiramezanali2019variational} or hyperbolic space representations \cite{sun2021hyperbolic}.

\textbf{BGNNs.}
A series of BGNN works consider graph structures, model parameters, and passed messages in graph learning as distributions rather than deterministic values. 
Bayesian GCNN \cite{zhang2019bayesian} treats the observed graph $G_{obs}$ as a sample from an assortative mixed membership stochastic block model (a-MMSBM) ~\cite{li2016scalable,gopalan2012scalable}. 
The posterior probability of node (or graph) labels can be computed by:
\begin{align}
    \label{eq: BGCN_random_graph_post}
    P(Y^*|Y_{\mathcal{T}}, \mathbf{X}, G_{obs})=&\int_{\Theta}\int_{\mathcal{G}}\int_{\Lambda} P(Z|\theta, G, \mathbf{X})P(\theta|Y_{\mathcal{T}}, G, \mathbf{X})\nonumber \\ 
    &\cdot P(G|\lambda)P(\lambda|G_{obs})\,d\theta\,dG \,d\lambda\,,
\end{align}
where the integrals are taken over the parameter space $\Theta$, the graph space $\mathcal{G}$, and the hyperparameter space $\Lambda$.
Here $\theta$ is a random variable representing the weight matrices of a Bayesian GCN over graph $G$, and $\lambda$ denotes the 
hyperparameters governing the prior distribution over the graph topology.
$Y_{\mathcal{T}}$ represents the training label set.
$P(Y^*|\theta, G, \mathbf{X})$ can be modeled using a categorical distribution by applying a softmax function to the final-layer node embeddings from the $L$-layer GCN. 
Alternative generative models for graph structure inference include node copying \cite{pal2019bayesian} which explicitly models the graph generation process, 
and non-parametric approaches \cite{pal2020non} that construct posterior distributions over adjacency matrices without restrictive parametric assumptions.




Beyond modeling graph structures as random variables, several works place distributions over other GNN components: BUP \cite{xu2022uncertainty} models messages as multivariate Gaussians with uncertainty-aware covariance propagation, GKDE \cite{zhao2020uncertainty} treats parameters as random variables to decompose EU and AU while incorporating evidence-theoretic measures of vacuity (reflecting a lack of evidence, corresponding to EU) and dissonance (reflecting conflicting evidence, corresponding to AU), and Elinas et al. \cite{elinas2020variational} assign Bernoulli priors to adjacency matrix entries with variational posterior inference.

For explicit AU-EU decomposition, BGNN \cite{munikoti2023general} propagates AU through variance tracking while estimating EU via MC dropout, whereas Fuchsgruber et al. \cite{fuchsgruber2024uncertainty} define EU as the information gain from label acquisition through entropy-based decomposition, demonstrating its effectiveness in guiding graph active learning.

\underline{\textit{Summary}}. 
Bayesian representation learning for graphs assigns probability distributions over latent variables or model components to represent uncertainty. GVAEs model stochastic embeddings of nodes or graphs, capturing uncertainty in learned representations, while BGNNs place distributions over graphs, messages, and parameters, capturing uncertainty in model components. However, the lack of closed-form posteriors requires approximate inference, which can be computationally intensive and may introduce additional errors.

\subsection{Gaussian Process}
\label{subsec: gaussain process}
\subsubsection{Gaussian Process Overview}\hfill\\
\indent Gaussian Processes (GPs) are a popular class of non-parametric Bayesian models widely used for UQ \cite{williams2006gaussian}. 
A GP is defined as a collection of random variables, any finite number of which have a joint Gaussian distribution. In the context of function approximation. 
GP can be used as a prior distribution over the space of functions, denoted as:

\begin{equation}
f(x) \sim \text{GP}(m(x), k(x, x')),
\end{equation}
where the mean function $m(x)$ calculates the expectation of the function at input $x$.
The covariance (kernel) function $k(x, x^\prime)$ encodes the correlation between function values at different inputs and determines the smoothness of the functions sampled from the GP.

%

Given a labeled dataset, 
the mean and covariance of the posterior and predictive distributions admit closed-form solutions 
directly yielding UQ; however, standard inference requires $\mathcal{O}(n^3)$ matrix inversion, so practical systems rely on sparse/approximate methods \cite{quinonero2005unifying, titsias2009variational, rahimi2007random}.
Recent advancements have also explored the integration of GPs with deep learning architectures to leverage the strengths of both approaches \cite{damianou2013deep, bradshaw2017adversarial}.

\subsubsection{Gaussian Process for Graphs}\hfill\\
\indent Applying GPs for graphs offers new possibilities for uncertainty estimation in graph learning tasks. 
GGP \cite{ng2018bayesian} defines a GP prior over node embeddings, using a kernel function that can be selected from existing well-studied options. 
It combines the GP prior with a robust-max likelihood \cite{girolami2006variational} to handle multi-class problems. 
As an extension of GGP for semi-supervised node classification, 
UaGGP \cite{liu2020uncertainty} addresses uncertainty arising from noisy or unreliable graph structures, where the presence or absence of edges may be uncertain, thereby affecting prediction confidence.
UaGGP uses GP to guide the model learning by incorporating prediction uncertainty and label smoothness regularization. 
For any two nodes $V_i$ and $V_j$, UaGGP defines an uncertainty-aware distance measure $\text{dist}(V_i, V_j)$ to constrain the proximity of the predictions at those nodes. 
This mechanism manifests UQ by ensuring that $\text{dist}(V_i, V_j)$ increases whenever the model's confidence in predictions is low.

Subsequent extensions address scalability and specialized tasks: DGPG \cite{li2020stochastic} proposes stochastic deep GPs for scalable node regression, 
while LNK \cite{wollschlager2023uncertainty} combines GNN-based encoders with sparse variational GPs to efficiently scale uncertainty estimation by mitigating the computational bottleneck of standard GPs.
FT-GP and WT-GP \cite{opolka2020graph} employ Fourier and wavelet transforms to capture multiscale and localized spectral information for graph classification. Similarly, GPG \cite{venkitaraman2020gaussian} integrates the graph Laplacian into the GP covariance matrix to enhance spectral regularization of target signals. 
Beyond scalability and specialization, GGPN \cite{chen2022multi} uses GPs to learn distributed embedding functions rather than deterministic embeddings, enhancing flexibility in representing diverse graph structures and their inherent uncertainty.

\underline{\textit{Summary}}.
Gaussian processes for graphs extend nonparametric Bayesian methods to graph-structured data by defining GP priors over node embeddings with graph-aware kernel functions. A key advantage of GPs is their ability to provide closed-form posterior distributions for standard regression tasks, enabling exact UQ without requiring approximate inference. 
While GPs are not strictly limited to regression and can be naturally adapted for classification, doing so involves non-Gaussian likelihoods that render exact closed-form solutions intractable, thus necessitating approximation techniques.
Subsequent extensions address scalability through sparse approximations, incorporate spectral information via transforms, or learn flexible kernel structures to capture diverse graph properties. These approaches provide principled UQ grounded in Bayesian theory, but face computational challenges in large-scale graphs and require careful kernel design to effectively encode graph topology.

\subsection{Stochastic Differential Equations}
\label{sec:SDEs_representation}

Stochastic Differential Equations (SDEs) offer a dynamic systems perspective for modeling uncertainty in neural networks~\cite{oksendal2003stochastic, li2020scalable}. Instead of viewing transformations as deterministic functions, SDE-based approaches model the evolution of hidden states or network parameters as stochastic processes. This continuous-time formulation provides a natural way to incorporate and quantify uncertainty throughout the network.

\subsubsection{SDEs Overview} \hfill \\
\indent An SDE for a latent state process \(Z_t\) is defined as:
\begin{equation}
dZ_t = f(Z_t, t)dt + g(Z_t, t)dW_t
\label{eq:general_sde}.
\end{equation}
The drift function \(f(Z_t, t)\) dictates the deterministic part of the evolution, and the diffusion function \(g(Z_t, t)\) scales the stochastic part driven by a Wiener process (Brownian motion) \(dW_t\)~\cite{oksendal2003stochastic}. Both can be parameterized by neural networks.
The core idea behind using SDEs for UQ is to leverage the stochastic diffusion term to transform deterministic forward passes into a distribution of possible trajectories; measuring the variance across these trajectories naturally quantifies the model's predictive uncertainty ~\cite{oksendal2003stochastic, karatzas1991brownian}.

SDE-Net \cite{kong2020sde} instantiates the general SDE in Eq. \eqref{eq:general_sde} by two neural networks, namely ``drift net'' and ``diffusion net''. 
The diffusion net induces a distribution over the final hidden state $Z_T$.  
Following the decomposition proposed in \cite{kong2020sde}, the accumulated stochasticity from the diffusion process results in a variance $\mathrm{Var}(Z_T)$ that reflects the model's uncertainty in representation, thereby quantifying epistemic uncertainty.  
AU is then obtained from an independent output layer applied to the terminal state $Z_T$ to form the predictive distribution $P(Y\mid Z_T)$, which models the inherent data noise given the representation.
%
Beyond SDE-Net, SDEs are employed to conceptualize infinitely deep BNNs, allowing scalable approximate inference by leveraging a continuous-depth perspective \cite{xu2022infinitely}. 
Another line of work formulates UQ within stochastic optimal control problems, using backward SDEs for gradient computation and uncertainty estimation \cite{archibald2022backward}. 
For time-series data, SDE-RNN \cite{dahale2023general} proposes a framework to quantify and propagate both AU and EU within recurrent architectures.



\subsubsection{SDEs for Graphs} \hfill \\
\indent SDEs are increasingly employed UQ on GNN by modeling the evolution of graph representations as continuous-time stochastic processes. 
LGNSDEs~\cite{bergna2025uncertainty} instantiates the general SDE in Eq.~\eqref{eq:general_sde} for graph data by parameterizing both drift and diffusion functions with GCNs, where learnable weights capture graph-specific dynamics to produce uncertainty-aware node representations.

GNSD \cite{lin2024graph} connects GNNs with Stochastic Partial Differential Equations (SPDEs) by modeling the evolution of node representations using graph operators such as Laplacians or neighborhood attention in the drift and diffusion terms.
For EU quantification, it employs a stochastic forcing network with a Q-Wiener process to model uncertainty propagation across the graph, contrasting with LGNSDE's Bayesian approach.
Additionally, AGGDN~\cite{xing2023aggdn} proposes a hybrid architecture that stacks an SDE module upon an ODE component, where the ODE provides deterministic control signals to guide the stochastic SDE propagation.
SDEs also extend to dynamic graphs and generative modeling. For dynamic graphs, SDEs are used in decoder architectures to infer causal graph structures in spatio-temporal forecasting~\cite{liang2024dynamic}. For graph generation, continuous-time diffusion processes based on SDEs enable permutation-invariant sampling~\cite{huang2022graphgdp}.

\underline{\textit{Summary}}.
Stochastic Differential Equations model the evolution of graph representations as continuous-time stochastic processes, where drift terms capture deterministic dynamics through GNN architectures and diffusion terms quantify uncertainty. 
This framework naturally extends to dynamic graphs and generative modeling by parameterizing temporal evolution with graph operators (such as Laplacians or aggregation functions).  
However, uncertainty quantified through diffusion terms depends critically on the choice of stochastic processes and lacks distribution-free guarantees, making it sensitive to model misspecification and difficult to interpret compared to methods with explicit probabilistic semantics, such as Bayesian Neural Networks or Gaussian Processes.

\subsection{OOD Epistemic Uncertainty Representation}
\label{sec:outlier_representation}
Enhancing generalization is an effective strategy for improving models' performance under uncertainty.
It enables models to make accurate predictions for ``unexpected'' and previously unseen samples. 
These outlier samples, which significantly differ from the training (in-distribution, ID) data, are commonly referred to as out-of-distribution (OOD) samples.
OOD samples typically correspond to high epistemic uncertainty: when a test input lies outside the support of the training distribution, the model has limited knowledge about how to generalize, and its EU increases accordingly. Therefore, effective UQ should assign high EU to truly novel regions while remaining calibrated on ID data. 
In this subsection, we survey various approaches for modeling perturbations and representing outliers.

\subsubsection{Dataset-level uncertainty estimation} \hfill \\
\label{sec:dro_representation}
\indent
One approach to represent OOD uncertainty is dataset-level uncertainty estimation \cite{qian2024harnessing}, 
where the distributional shifts are captured by defining an uncertainty set over data distributions. 
In this way, unseen data are modeled as elements within a specified neighborhood of the empirical training distribution. 
Let 
$\mathcal{P}:=\{ Q | \text{dist}(Q,P) \leq \rho\}$
define the uncertainty set, where $P$ is the observed data distribution, and $\rho$ denotes a radius distance.
The distance between two distributions, $Q$ and $P$, is measured by a divergence metric $\text{dist}(\cdot, \cdot)$.
We list the most widely used choices of $\text{dist}$.

\begin{itemize}
\item 
\underline{$f$-divergence},
also known as $\phi$-divergence, is defined by
$\text{dist}_f(Q||P) := \int f\left(\frac{d Q}{dP}\right)\,\dd{P}$.
It is widely used to quantify the divergence between two distributions \cite{duchi2021learning,bayraksan2015data,ben2013robust}.
The domain delineated by $f$-divergence can be viewed as a statistical confidence region, and the optimization problem 
could be 
tractable \cite{ben2013robust,pardo2018statistical} for several choices of $f$, such as \underline{KL-divergence}
\cite{hu2013kullback,zhu2022shift,fang2025homophily}.
However, 
$f$-divergence based metrics, including KL-divergence,
may not be comprehensive enough to include some relevant distributions, and they may fail to precisely measure the divergence for extreme distributions \cite{gao2023distributionally}.


\item 
\underline{Wasserstein distance} is a geometry-aware discrepancy that alleviates the above limitations \cite{mohajerin2018data,zhao2018data,kuhn2019wasserstein,gao2023distributionally}. 
Formally, it is defined by
$\text{dist}_{W}(Q,P) := \inf_{\gamma\in T(Q,P)} \int \int \gamma(q,p) c(q, p) \, \dd{q} \, \dd{p}$, 
where $T(Q,P)$ contains all possible couplings of $Q$ and $P$, 
and $c(q, p) \geq 0$ are some cost functions transferring from $q$ to $p$. 

\item 
\underline{Maximum mean discrepancy (MMD)} \cite{staib2019distributionally} and the \underline{Prohorov metric} \cite{erdougan2006ambiguous} 
provide alternative ways to define uncertainty sets. 
Both MMD and Wasserstein are integral probability metrics (IPMs), while MMD over the reproducing kernel Hilbert space, and Wasserstein over 1-Lipschitz functions via optimal transport duality.
However, MMD strongly depends on the kernel choice and bandwidth. 
Prohorov metric metrizes weak convergence and is broadly applicable, but it is typically less convenient computationally and often yields less tractable robust objectives.


\end{itemize}

In graph learning, Wasserstein distance has been applied to node classification and structure learning by defining costs over node features, edge attributes, or latent graph embeddings \cite{zhang2021robust,chen2021wasserstein,shi2024wasserstein,li2025model}.  
Moreover, 
Wu et al. \cite{wu2023non} introduce a novel graph subtree discrepancy based on the Weisfeiler–Lehman subtree kernel, comparing hierarchical structural patterns across graphs. 
Fang et al. \cite{fang2025homophily} define a graph-level heterophily distribution to characterize structural shifts induced by changes in homophily, thereby capturing uncertainty arising from relational rewiring across domains.


\underline{\textit{Summary}}.
Dataset-level uncertainty estimation captures epistemic uncertainty stemming from distributional shift between training and test data by positing an ambiguity set around the training distribution. 
$f$-divergences (e.g., KL) offer tractable formulations, while Wasserstein distance provides a geometry-aware notion of shift at higher computational cost. 
Alternatives, such as MMD and Prohorov, offer additional flexibility, and graph-specific discrepancy measures further adapt these ideas to relational domains.

\subsubsection{Sample-level uncertainty representation} \hfill \\
\label{sec: AT_representation}
\indent
Sample-level uncertainty refers to the uncertainty in the model's prediction for \textit{one} sample. 
\textit{Adversarial samples} are the prime illustration of such uncertainty. 
For example, 
Qian et al. \cite{qian2024harnessing} measure it by perturbing an input sample and observing how much the model prediction can be changed. 
 
By adopting adversarial attack methods such as FGSM \cite{goodfellow2014explaining} and PGD \cite{madry2017towards}, 
adversarial samples leading to high task-specific loss $\mathcal{L}$ can be generated.
These adversarial samples essentially highlight regions where the model is uncertain and sensitive to small changes, and thus serve as a practical proxy to quantify sample-level uncertainty. 
%
Formally, adversarial attacks can be defined:
\begin{equation}
    \epsilon^\ast = \arg \max_{\|\epsilon\|\leq \rho} \mathcal{L}(\theta,x+\epsilon,y),
\end{equation}
where $(x,y)$ denotes a sample $x$ with its corresponding label $y$. 
$x$ is manipulated by an adversarial perturbation $\epsilon$, constrained by a predefined small budget $\rho$.
$x+\epsilon^\ast$ is the adversarial counterpart of original input $x$.

\underline{\textit{Summary}}.
Sample-level uncertainty representation characterizes the model's sensitivity to perturbations in individual samples, reflecting local instability
of a prediction.
Adversarial perturbations effectively probe this sensitivity by identifying minimal perturbations that can induce misclassification. 
Through such perturbation-based analysis, regions of high local variance in model predictions are interpreted as areas of elevated epistemic uncertainty. 
Consequently, sample-level representation offers a fine-grained measure of epistemic uncertainty, complementing dataset-level methods by exposing vulnerabilities and improving robustness per sample.

\subsubsection{Manually designed uncertainty sets} \hfill \\
\label{sec: SSL_representation}
\indent
When imposing a \textit{formal} probability distribution on inputs or model parameters is hard, 
one can approximate the uncertainty by manually designed controllable perturbations to replace mathematical formulation. 
These perturbations are carefully designed to preserve the core semantic or structural content while introducing realistic variations. 
By observing how the model's predictions vary across these perturbations, we obtain a proxy measure of epistemic uncertainty.


Specifically, manually designed uncertainty sets can be instantiated by augmenting input graphs through controlled perturbations that retain salient structural or attributive information. 
By adopting designed perturbations, input graphs are augmented in a manually controllable way.
Structural augmentations include node/edge deletion or addition \cite{you2020graph}, and subgraph sampling via random walks to preserve local topology \cite{qiu2020gcc}. 
At the attribute level, node and edge features can be masked or corrupted to mimic missing or noisy information. Some works \cite{zhu2020deep} combine both structural and attribute perturbations to generate diverse graph augmentations.

\underline{\textit{Summary}}.
Manually designed uncertainty sets provide an interpretable and controllable framework for approximating EU when explicit probabilistic modeling is intractable. 
By constructing perturbations that maintain key semantic or topological properties, these methods explore the model's predictive stability under realistic input variations.

\begin{table*}[h]
\centering
\caption[Taxonomy for Uncertainty Handling Methods.]
{Summary of Uncertainty Handling studies on graphs.\footnotemark[2].}
\def\arraystretch{1.2}
  \begin{tabular*}{0.85\textwidth}{c|c|c|c|c|c|c|c}
    \toprule
\textbf{Type}&\textbf{Methods}&\textbf{Task}& \textbf{Post-hoc}&\textbf{Non-param.}& \textbf{\makecell{Input\\adaptive}} & \textbf{\makecell{Topology\\aware}} & \textbf{Note}\\
\midrule
\multirow{6}{*}{\makecell{Bayesian \\Inference\\
}}&
    GPN~\cite{stadler2021graph}        & Node Cls   &            &            & \checkmark &  \checkmark & Exact\\
    &VGAE~\cite{kipf2016variational}     & Node RL   &            &  & \checkmark &  \checkmark & Approximated\\
    &Bayesian GCNN~\cite{zhang2019bayesian}     & Node Cls   &            &  & \checkmark &  \checkmark & Approximated\\
    &GGP~\cite{ng2018bayesian}     & Node Cls   &            & \checkmark & \checkmark &  \checkmark & Approximated\\
    &LGNSDE~\cite{bergna2025uncertainty}     & Node Cls   &            &  & \checkmark &  \checkmark & Approximated\\
    &E-$\Delta$UQ~\cite{10445967}& Link Pred. & & &\checkmark & \checkmark &Approximated\\

\midrule

\multirow{5}{*}{\makecell{OOD\\Detection\\Techniques}}&
    DR-GNN \cite{wang2024distributionally} & Node Cls & & & & \checkmark & DRO \\
    &GraphDefense \cite{wang2019graphdefense} & Node Cls & & & \checkmark & & AT \\
    &GraphAT \cite{feng2019graph} & Node Cls & & & \checkmark & \checkmark & AT \\
    &GraphCL \cite{you2020graph} & Graph Cls & & \checkmark & & \checkmark & SSL \\
    &GRACE \cite{zhu2020deep} & Node Cls & & \checkmark & & \checkmark & SSL \\
    \midrule
\multirow{6}{*}{Calibration}&
    Histogram Binning~\cite{zadrozny2001obtaining} &Cls& \checkmark& \checkmark & & & Binning-based\\

    &Isotonic Regression~\cite{zadrozny2002transforming} &Cls& \checkmark&\checkmark & & & Binning-based\\

    &Platt Scaling~\cite{platt1999probabilistic}&Cls& \checkmark& & & & Logit-based \\



    &CaGCN~\cite{wang2021confident}& Node Cls &\checkmark&&\checkmark&\checkmark & Logit-based \\


    &RBS~\cite{liu2022calibration}& Node Cls &\checkmark& & &\checkmark& Logit-based \\ 


    &IN-N-OUT~\cite{zhang2018link}& Link Pred. &\checkmark&&\checkmark&\checkmark& Logit-based\\

    \midrule
    \multirow{6}{*}{\makecell{Conformal\\ Prediction}}
    
    &Inductive CP~\cite{papadopoulos2002inductive}& Regr \& Cls & \checkmark & \checkmark & & & Rely on i.i.d. \\


    &APS~\cite{romano2020classification}& Cls &\checkmark & \checkmark &\checkmark & & Rely on i.i.d.\\

    &NexCP~\cite{barber2023conformal}& Regr \& Cls & \checkmark & \checkmark & & & Beyond i.i.d. \\

    &DAPS~\cite{zargarbashi2023conformal} & Node Cls& \checkmark & \checkmark &\checkmark &\checkmark & Rely on i.i.d.\\
    
    &Jurygcn~\cite{kang2022jurygcn} & Node Cls& \checkmark & \checkmark & & & Beyond i.i.d.\\

    &NAPS~\cite{clarkson2023distribution} & Node Cls& \checkmark & \checkmark &\checkmark & \checkmark & Beyond i.i.d.\\
    
    \bottomrule
\end{tabular*}
\label{tab:uncertainty_handling}
\end{table*}

\section{Methods for Uncertainty Handling}
\label{sec:method_uncertainty_handling}

As shown in Table \ref{tab:uncertainty_handling},
we explore four key approaches for managing uncertainty.
These methods offer distinct strategies for addressing scenarios where models 
directly handle uncertainty, encounter data outside its training distribution, adjust model confidence, and ensure robust predictions.

\subsection{Bayesian Inference Techniques}\hfill\\
\label{subsec: bayesian_inference_techniques}
We first discuss exact Bayesian inference methods that compute posterior distributions directly from Bayes theorem. We then present approximate methods that use Monte Carlo dropout, Variational Inference, and anchoring.

\subsubsection{Exact Bayesian Inference} \hfill \\
\label{subsec: exact_bayesian_inference}
\noindent \textbf{Exact Bayesian Inference Overview.} Exact Bayesian inference refers to methods that derive closed-form posterior distributions by directly applying Eq. (\ref{eq:Bayes_Theorem}) to the parameters of interest.
This approach naturally fits node classification tasks, as neighboring nodes can be treated as observed data (or evidence) to infer a node’s posterior distribution of the parameters used for classification.

\noindent \textbf{Exact Bayesian Inference for Graphs.}  
To execute exact inference on graphs, methods must bypass the intractable integrals typically associated with marginal likelihoods. They achieve this primarily by exploiting conjugate priors, which guarantee that the posterior distribution belongs to the same probability family as the prior, thereby enabling precise, closed-form algebraic updates.

In classification tasks, the inference process frequently relies on Dirichlet-Multinomial conjugacy. Rather than utilizing sampling or variational approximations, uncertainty is handled by directly aggregating network evidence into additive pseudo-counts. Specifically, SocNL \cite{yamaguchi2015socnl} executes inference by algebraically combining neighbor labels (treated as multinomial observations) with a Dirichlet prior to obtain per-node posteriors in closed form; 
NETCONF \cite{eswaran2017power} adjusts this inference mechanism by using a compatibility matrix to propagate label evidence before applying the analytical conjugate update; 
and GPN \cite{stadler2021graph} computes the exact posterior by mapping features to Dirichlet pseudo-counts with a learned transformation and diffusing them over the graph prior to the final conjugate updating. 
Collectively, these approaches exemplify how relying on conjugate analytical solutions allows graph models to handle uncertainty through exact, deterministic calculations rather than stochastic approximations.

\noindent \textbf{Exact Bayesian Inference for GFMs.} 
To our knowledge, exact Bayesian inference for GFMs remains largely unexplored. Few techniques have been proposed in foundation models like LLMs, which provide valuable insights and references for GFMs. Lu et al.~\cite{lu2024uncertainty} utilize a Gaussian process classification layer by treating the pre-trained foundation model as a deterministic feature extractor to quantify uncertainty, and derive the closed-form expression for the Gaussian posterior of the class-specific logits for any new instance.

\underline{\textit{Summary}}.
Exact Bayesian inference for graphs leverages prior-likelihood conjugacy to compute closed-form posteriors, eliminating approximation errors in variational or sampling methods. By converting neighborhood evidence into additive counts, Dirichlet-based methods enable direct uncertainty quantification without iterative optimization. However, the requirement of conjugacy restricts applicability to specific model families and may not accommodate the flexible architectures needed for complex graph learning tasks. For GFMs, exact Bayesian inference remains unavailable, though techniques from LLMs offer potential pathways for future development.

\subsubsection{Approximate Bayesian Inference} \hfill \\
\label{subsec: approximate_bayesian_inference}
\noindent \textbf{Approximate Bayesian Inference Overview.} 
Approximate Bayesian inference is a well-established tool for handling uncertainty when exact Bayesian methods are computationally intractable. 
We review 
three key techniques: Monte Carlo (MC) Dropout \cite{gal2016dropout}, Variational Inference (VI) \cite{jordan1999introduction}, and Anchored Ensembling \cite{pearce2020uncertainty, thiagarajan2022single}. 

MC Dropout approximates the Bayesian posterior by interpreting the dropout mechanism as a variational approximation over the model weights. Specifically, given an input \( x \), the predictive distribution is estimated by averaging over multiple stochastic forward passes as follows:
\begin{equation}
\label{eq: MC_Dropout}
P(Y \mid x) \approx \frac{1}{T} \sum_{t=1}^{T} P\bigl(Y \mid x, \hat{\theta}^{(t)}\bigr),  
\end{equation}
where \( \hat{\theta}^{(t)} \) represents sampled model weights induced by applying dropout at each forward pass, and \( T \) is the total number of Monte Carlo runs.

In contrast, VI approximates the true posterior \( P(\theta \mid \mathcal{D}) \) by a tractable variational distribution \( Q_{\phi}(\theta) \), parameterized by \( \phi \). This approximation is achieved by maximizing the ELBO:
\begin{equation}
\label{eq:VI}
\mathcal{L}(\phi) = \mathbb{E}_{Q_{\phi}(\theta)}\Bigl[\log P(\mathcal{D}\mid \theta)\Bigr] - \text{dist}_\mathrm{KL}\Bigl(Q_{\phi}(\theta) \parallel P(\theta)\Bigr),
\end{equation}
which balances the data likelihood with a regularization term expressed as the 
KL-divergence between the variational distribution and the prior \( P(\theta) \). 

Beyond MC Dropout and VI, Anchored Ensembling offers a scalable alternative for approximate Bayesian inference. 
Pearce et al. \cite{pearce2020uncertainty} demonstrate that regularizing parameters towards randomized anchor points effectively approximates the Bayesian posterior around local modes. 
Building on this, $\Delta$-UQ \cite{thiagarajan2022single} introduces Stochastic Data Centering, a data-centric adaptation that perturbs the input space relative to random anchors rather than modifying model parameters. 
By marginalizing predictions over multiple anchors during inference, this approach efficiently estimates epistemic uncertainty without requiring complex variational objectives. 
Specifically, it quantifies the model's lack of knowledge by measuring how its predictive behavior fluctuates across different randomized reference points.

\noindent \textbf{Approximate Bayesian Inference for GNNs.}
Models specifically designed for uncertainty quantification in graph learning often employ approximate Bayesian inference techniques to handle uncertainty over model parameters, latent representations, and even the graph structure itself. In many cases, these techniques follow the same principles as those outlined in Eqs.~\eqref{eq: MC_Dropout} and \eqref{eq:VI}, but they are adapted to account for graph data \cite{pal2019bayesian, pal2020non, zhao2020uncertainty, munikoti2023general, bergna2025uncertainty}.

Bayesian GCNNs~\cite{zhang2019bayesian} leverage MC dropout to approximate the posterior distribution over both the model weights and the observed graph topology, as given in Eq.~\eqref{eq: BGCN_random_graph_post}.
\begin{equation}
\label{eq:BGCN_MC}
P(Y^* \mid Y_{\mathcal{T}}, \mathbf{X}, G_{obs}) \approx \frac{1}{N_G S} \sum_{i=1}^{N_G} \sum_{s=1}^{S} P\bigl(Y^* \mid \hat{\theta}_{s,i}, G_i, \mathbf{X}\bigr),
\end{equation}
where \(N_G\) is the number of graph samples, \(S\) is the number of MC dropout samples, \(G_i\) denotes a graph sample drawn from an assortative mixed membership stochastic block model (e.g., an a-MMSBM~\cite{li2016scalable,gopalan2012scalable}), and \(\hat{\theta}_{s,i}\) is a stochastic sample of weights (obtained via MC dropout). 

Variational Inference (VI) is also applied in graph settings to approximate the posterior over 
latent variables $\mathbf{Z}$.
For instance, VGAE \cite{kipf2016variational} obtains the solution for Eq.~\eqref{eq:VGAE} by optimizing the ELBO
\begin{equation}
    \label{eq:GVAE_ELBO}
    \mathbb{E}_{Q_{\phi}(\mathbf{Z}|\mathbf{X},\mathbf{A})}[\log\,P(\mathbf{A}|\mathbf{Z})] - \text{dist}_\mathrm{KL}\Bigl[Q_{\phi}(\mathbf{Z}|\mathbf{X},\mathbf{A}) \,\|\, P(\mathbf{Z})\Bigr].
\end{equation}
$\log P(\mathbf{A}|\mathbf{Z})$ is the reconstruction term, which encourages the model to accurately reconstruct the graph's adjacency matrix. 
The KL divergence term
acts as a regularizer, ensuring that the learned distribution does not deviate too far from the prior $P(\mathbf{Z})$.
Other works that similarly use VI to estimate the predictive posterior include approaches for node representations \cite{hasanzadeh2019semi, li2020dirichlet, ahn2021variational, grover2019graphite, hajiramezanali2019variational, sun2021hyperbolic, ng2018bayesian, liu2020uncertainty, wollschlager2023uncertainty, opolka2022adaptive, li2020stochastic, opolka2023graph, opolka2020graph, bergna2025uncertainty, liang2024dynamic}, graph representations \cite{simonovsky2018graphvae}, and graph structure \cite{elinas2020variational}. 

Complementing these approaches that approximate distributions over model parameters, stochastic data centering offers a data-centric route to epistemic uncertainty estimation. 
In the  graph domain, E-$\Delta$UQ~\cite{10445967} adapts this framework for link prediction by anchoring node features to quantify the uncertainty of edge existence. 
G-$\Delta$UQ~\cite{trivedi2024accurate} further extends the approach to graph-level classification through hierarchical anchoring strategies applied at input, intermediate, or readout layers. 
These methods provide a scalable alternative to ensembles by directly estimating uncertainty from input perturbations.

\noindent \textbf{Approximate Bayesian Inference for GFMs.} 
GFM uncertainty can be approximated by Bayesian ensembles, which are Monte Carlo approximation of the posterior distributions. Pasini et al. \cite{lupo2025scalable} build a trustworthy GFM via ensemble epistemic UQ for atomistic materials modeling. Sun et al.~\cite{sun2022quantifying} quantify uncertainty through evaluating ensemble disagreement, computing mean and standard deviation of the likelihoods for the same samples in different foundation models. BayesPE~\cite{tonolini2024bayesian} quantifies uncertainty in LLMs by computing output probabilities through ensemble prompts, effectively approximating a Bayesian input layer and providing a lower bound on the expected error. BLoB~\cite{wang2024blob} also introduces a principled Bayesian framework for Low-Rank Adaptation (LoRA) in LLMs, and assumes that posterior distribution of the weights are linear combination of independent Gaussian distributions. Those approaches for foundation models provide potential insights for GFMs. 

\underline{\textit{Summary}}.
Approximate Bayesian inference addresses intractable posteriors in GNNs and GFMs by trading exact computation for scalable estimation. These methods enable Bayesian reasoning in graph architectures where conjugacy or analytical solutions are unavailable. 
However, the gap between approximate and true posteriors is generally intractable to compute, and convergence guarantees may be weak, potentially leading to unreliable estimates in practice.
\subsection{Handling uncertainty due to OOD data}
\label{sec:ood_handling}
When a model encounters OOD data, its \textit{epistemic} uncertainty on those data points is typically higher.
To handle the uncertainty, a surge of research has focused on OOD generalization,
and most of them propose novel network architectures and objective functions to improve the model performance on OOD test data.
We organize the discussion and review the studies according to the taxonomy of OOD handling strategies \cite{li2022out},
which categories the methods into 
distributionally robust optimization (DRO) 
\cite{duchi2021learning,bayraksan2015data,ben2013robust,pardo2018statistical,hu2013kullback,gao2023distributionally,mohajerin2018data,zhao2018data,kuhn2019wasserstein,staib2019distributionally,erdougan2006ambiguous,wang2022distributionally,sadeghi2021distributionally,chen2023uncertainty,chen2021wasserstein,shi2024wasserstein,li2025model}, 
adversarial training (AT) \cite{wu2023adversarial,xue2021cap,gosch2024adversarial,xu2019topology,chen2019can,geisler2021robustness,wang2019graphdefense,deng2023batch,feng2019graph}, 
and self-supervised learning (SSL) \cite{jaiswal2020survey,liu2022graph,pathak2016context,zhang2016colorful,noroozi2016unsupervised,chen2020simple,he2020momentum,mikolov2013efficient,mikolov2013distributed,devlin2018bert,you2020graph,you2020does,hu2019strategies,zhu2020deep,yehudai2021local,liu2022confidence}.

\subsubsection{Distributionally Robust Optimization} \hfill \\
\label{sec: DRO_handling}
\noindent \textbf{Distributionally Robust Optimization Overview.}
Distributionally Robust Optimization (DRO) aims to enhance the model's performance across a wide range of potential testing data distributions \cite{duchi2021statistics,ben2013robust}.
Specifically, DRO optimizes the objective function in the ``worst case'' of sample distributions $P \in \mathcal{P}$.
It is formulated as a bi-level optimization problem: 
\begin{equation}
\label{eq:dro_objective}
    \min_{\theta} 
    \max_{P\in\mathcal{P}} \mathbb{E}_{(x,y)\sim P}[\mathcal{L}(\theta, x, y)], 
\end{equation}
where $\mathcal{L}$ is a loss function.
The uncertainty set $\mathcal{P}$, formally defined as $\mathcal{P}:=\{ P | \text{dist}(Q_{XY},P_{XY}) \leq \rho\}$, encapsulates all the unseen data distributions within a specific distance from the training distribution.
As discussed in Sec. \ref{sec:dro_representation}, several divergence metrics $\text{dist}(\cdot,\cdot)$ have been adopted in DRO.
\noindent \textbf{Distributionally Robust Optimization for GNNs.}
Due to the unique topological structure of the graphs,
the distance metric should measure the discrepancy between distributions of node representations \cite{wang2024distributionally,sadeghi2021distributionally,wang2022distributionally}, graph representations \cite{wu2024understanding}, or labels \cite{chen2023uncertainty}.

One widely used distance metric in DRO for graphs is the Wasserstein distance \cite{zhang2021robust,chen2023uncertainty,wang2022distributionally}.
Specifically,
Zhang, et al. \cite{zhang2021robust}
propose a graph learning framework that instantiates DRO under two uncertainty assumptions: one variant models distributions within the uncertainty set using a Gaussian family, while the other adopts a nonparametric formulation without distributional priors. 
Beyond Wasserstein neighborhoods, Wang et al. \cite{wang2022distributionally} develop a moment-based DRO model for learning a graph Laplacian from noisy smooth signals. The key observation is that the expected smoothness risk is determined by the first two moments of the signal distribution, enabling an ambiguity set defined by neighborhoods around the empirical mean and covariance. It avoids estimating a full distribution, and thus works well when only limited samples are available, and it can yield more stable out-of-sample behavior by optimizing against worst-case distributions consistent with observed low-order statistics. 
Assuming that nodes belonging to the same class follow the same underlying feature distribution, 
the uncertainty set includes an underlying distribution of node embeddings (based on Wasserstein distance) for each class \cite{chen2023uncertainty}.

Furthermore, some works \cite{wang2024distributionally,wu2024understanding} explore the connection between DRO (using the KL-divergence as the distance metric) and graph learning methods. 
Specifically, 
Wu et al. \cite{wu2024understanding}
reveal that \textit{contrastive learning} implicitly performs DRO over the negative sampling distribution.
Wang et al. \cite{wang2024distributionally} prove that 
LightGCN
\cite{he2020lightgcn} works as a graph smoothness regularizer within the DRO framework, 
enabling a regularization-based method that promotes robustness and generalization.

\noindent \textbf{Distributionally Robust Optimization for GFMs.} GFMs pretrained on large-scale graph corpora can learn generalizable patterns, but cannot inherently ensure robustness to OOD data. To improve the robustness, GLIP-OOD \cite{xu2025glip}, LLM-GOOD \cite{xu2025few} and GOE-LLM \cite{xu2025graph} leverage LLMs to check whether node texts match known in-distribution category, and utilize GFMs to perform zero-shot OOD detection based on node texts. Some studies discuss the distributional uncertainty arising from structures. Wang et al. \cite{wang2025text} fuse local structures into node-level text representations, and generates node-specific transformation to enhance OOD distinction across structural shifts. MDGFM \cite{wang2025multi} bridges domains by balancing features and topologies, and refining original graphs to eliminate noise and align structures. Textual-attributed GFM UltraTAG \cite{zhang2025toward} follows the homophily principle to reconstruct missing textual attributes from neighbors using LLMs, and generates contextually relevant texts to reduce uncertainty.

\underline{\textit{Summary}}.
DRO optimizes model performance under the worst-case distribution within an uncertainty set, and provides theoretically grounded guarantees for robustness against unseen environments. 
Recent advances in graph learning extend DRO to node and graph levels, where uncertainty sets capture discrepancies in feature embeddings, label distributions, or graph structures. 
DRO 
bridges statistical robustness 
and graph-specific uncertainty modeling.
With GFMs, DRO further enhances robustness beyond large-scale pre-training, and enables systematic handling of semantic and structural distribution shifts encountered in OOD graph data.

\subsubsection{Adversarial Training} \hfill \\
\label{sec: AT_handling}
\noindent \textbf{Adversarial Training Overview.}
Adversarial Training (AT) generates adversarial samples, which degrade model performance most with imperceptible perturbations \cite{madry2017towards}.
The models will then be trained with these adversarial samples to improve robustness. 
Formally,
AT solves the following optimization problem:
\begin{equation}
\label{eq:at_overview}
    \min_{\theta} \max_{\|\epsilon\|\leq \rho} \mathcal{L}(\theta, x+\epsilon, y).
\end{equation}
\noindent \textbf{Adversarial Training for GNNs.} 
%
Considering both the attributional and topological features of graphs, 
AT for graphs considers the following min-max optimization problem: 
\begin{equation}
\label{eq:at_graph}
    \min_{\theta} 
    \max_{\epsilon_{\theta}, \epsilon_{x}, \epsilon_{a}} 
    \mathcal{L}(\theta + \epsilon_{\theta}, \mathbf{A}+\epsilon_{a}, \mathbf{X}+\epsilon_{x}, Y),
\end{equation}
where $\epsilon_{\theta}$, $\epsilon_{a}$, and $\epsilon_{x}$ represent the perturbations applied to model parameters $\theta$, adjacency matrix $\mathbf{A}$, and node feature matrix $\mathbf{X}$, respectively.
These perturbations are constrained by a budget, e.g., $\text{dist}(\mathbf{X}+\epsilon_x, \mathbf{X}) \leq \rho_x$.
Consequently, the uncertainties from model parameters, graph topologies, and node features are incorporated during the generation of adversarial samples. 
We summarize the existing studies from these three aspects.

A branch of AT studies on graphs focuses on manipulating \textit{model parameters} $\theta$ to improve generalization \cite{wu2023adversarial,xue2021cap},
where the perturbations are evaluated by $L_p$ norm.
Some studies \cite{gosch2024adversarial,xu2019topology,chen2019can,geisler2021robustness,wang2019graphdefense} focus on manipulating \textit{graph topologies}, such as adding or deleting edges, 
corresponding to $\epsilon_a$ in Eq. (\ref{eq:at_graph}). 
For the distance metric $\text{dist}$,
many works \cite{gosch2024adversarial,xu2019topology,chen2019can,geisler2021robustness} count the number of malicious edge modifications by the $L_0$ norm.
In other work \cite{wang2019graphdefense}, discrete edge modifications are relaxed to continuous changes,
evaluated using the $L_1$ norm.
%
Perturbing node features is also a commonly explored strategy, referring to $\epsilon_x$ in Eq. (\ref{eq:at_graph}),
where the perturbations are usually constrained by $L_p$ norms (e.g., $L_1$, $L_2$, or $L_\infty$) \cite{dai2019adversarial,deng2023batch,feng2019graph} and the Wasserstein distance \cite{wang2023toward}.

\noindent \textbf{Adversarial Training for GFMs.} While recent studies have investigated adversarial textual and structural attacks against GFMs like LLM-as-Predictors~\cite{guo2024learning} and LLM4RGNN~\cite{zhang2025can}, limited attention has been given to AT specifically targeting topological graph features. GraphCLIP~\cite{zhu2025graphclip} exposes the model to perturbed graph structures, and enforces alignment invariance between structural and semantic representations. GRAVER~\cite{yuan2025graver} handles uncertainty by augmenting the few-shot support set with generative perturbations, which acts as a semantic and structural perturbation generation mechanism that stabilizes cross-domain fine-tuning.

\underline{\textit{Summary}}.
AT enhances GNN's robustness by explicitly exposing models to worst-case perturbations in a min-max optimixzation problem. 
Unlike DRO, which defines uncertainty sets via distances between \emph{data distributions},
AT relies on $L_p$ norm as distance metrics to specify a bounded neighborhood of allowable variations.
$L_p$ norm is \emph{not} an uncertainty metric and does not quantify EU.
However, if a model’s predictions vary substantially within the $L_p$-bounded neighborhood, this variability can be interpreted as evidence of limited model knowledge around the input. 
AT integrates uncertainty handling into the learning process, offering resilience against adversarial and naturally occurring distributional shifts, and serving as a practical complement to the theoretical robustness guarantees provided by DRO.

\subsubsection{Self-Supervised Learning} \hfill\\
\noindent \textbf{Self-Supervised Learning Overview.}
Self-Supervised Learning (SSL) allows models to learn informative signals from input samples and their variants, reducing the demands for labeled samples \cite{jaiswal2020survey,liu2022graph}.
SSL typically designs pretext tasks to extract meaningful features and enable models to learn generalized representations, providing a good initialization for downstream tasks.
For example, 
SSL on computer vision could adopt
image inpainting, where 
models learn to predict the missing parts masked by design \cite{pathak2016context}, as a pretext task. 


\noindent \textbf{Self-Supervised Learning for GNNs.}
SSL on graphs involves deliberately perturbing graphs to construct new controllable datasets while preserving the intrinsic information of the original graphs. 
These augmented variations of the graph instances allow SSL to simulate real-world shifts in data distributions
and allow models to better handle EU by training on these augmented graphs. 
We categorize SSL methods based on \textit{structural} and \textit{attributional} feature perturbations.

Some studies focus on altering the \textit{structural} features of graphs.
For instance, 
You et al. \cite{you2020graph} 
propose augmentations where a portion of nodes is randomly removed, and a set of edges can be either deleted or added. 
These modifications generate new graph instances while largely preserving global connectivity patterns. 
Besides, they propose subgraph extraction using random walk.
The subgraph extraction method is also employed to capture local structure effectively \cite{qiu2020gcc}.

Other studies propose to manually design modification for the \textit{attributional} features of graphs. 
Specifically,
inspired by image inpainting \cite{pathak2016context},
graph completion is proposed in \cite{you2020does}, where the attributional features of nodes are masked, and a model learns to recover these missing features.
Similarly, 
Hu et al. \cite{hu2019strategies}
propose randomly masking attributional features on nodes and/or edges to create variations in the graph without changing its overall structure.
Furthermore, 
You et al. \cite{you2020does}
propose node clustering and graph partitioning methods,
where nodes are clustered based on feature similarity or topological connections, and cluster indices are used as pseudo-labels for SSL tasks.
Zhu et. al \cite{zhu2020deep} augment graph data at both the structure and attribute levels, providing a more comprehensive approach to generating new graphs.

Beyond prediction, 
some studies adopt SSL augmentation for distinct goals.
For example, 
Yehudai et al. \cite{yehudai2021local}
propose a novel SSL task centered on predicting the $d$-pattern tree descriptor, 
encouraging representations to encode local topological patterns.
Meanwhile, 
Liu et al. \cite{liu2022confidence}
address the distribution shift issue caused by high-confidence unlabeled nodes. 
They propose a collaborative framework based on information gain to reweight samples, thereby mitigating the effects of this shift.
This work illustrate that SSL can be used not only for representation learning but also to improve robustness under distribution shift for OOD cases.

\noindent \textbf{Self-Supervised Learning for GFMs.} Recent studies employ self-supervised objectives to learn highly transferable structural and feature representations. RiemannGFM \cite{sun2025riemanngfm} assumes that shared structural knowledge for cross graph domains can be learned on a Riemannian manifold, and pretrains a structurally universal self-supervised GFM inherently from Riemannian geometry. LLM-based GFMs leverage LLMs as node attribution taggers, OGA \cite{wen2025llms} designs a graph label annotator via LLMs, annotating unknown classes from node concepts and semantic in-context information. Inspired by the token vocabulary in LLMs, REEF \cite{yu2025relation} leverages relation tokens as the basic units for GFMs, enabling effective transfer across different graph domains via self-supervised relational tasks. 

\underline{\textit{Summary}}.
SSL leverages self-generated supervision, such as pretext tasks and controllable graph augmentations, to learn invariant and transferable representations that remain stable under structural or attributional perturbations. 
While these augmentations help simulate different data variations and implicitly introduce uncertainty, they do not explicitly measure or quantify the EU caused by these transformations. 
Nonetheless, SSL augmentations are often more diverse and challenging due to significant perturbations over inputs. 
\subsection{Calibration}
\label{sec:calibration}

\noindent \textbf{Calibration Overview.}
We consider a classifier $f$ with label space $\mathcal{Y} = \{1,\ldots,K\}$. 
Given an input $x$ and its true label $y$, the classifier applies a softmax function 
to produce a probability vector $(\pi^{(1)},\ldots,\pi^{(K)})$, where $\pi^{(k)}$ 
denotes the predicted probability of class $k$ and satisfies
$\sum_{k=1}^K \pi^{(k)} = 1$. The predicted label $\hat{y}$ and and confidence $\hat{p}$ are given by
\begin{equation}
    \hat{y}=\arg \max_{k\in \mathcal{Y}} \pi^{(k)},\text{ } \hat{p}=\max_{k\in \mathcal{Y}} \pi^{(k)}.
\end{equation}
A model is considered well-calibrated if the confidence matches the probability of correct prediction:  $\text{Pr}(\hat{y}={y}|\hat{p}=c)=c, \forall c \in [0,1]$. 
A stronger definition of being well-calibrated requires that each class (not only the predicted label) is calibrated:
$\text{Pr}(y=k|\pi^{(k)}=c)=c, \text{ } \forall k \in \mathcal{Y}, \text{ } \forall c \in [0,1]$. 


%
Calibration can be conducted on top of a trained, accuracy-driven model to adjust prediction confidences via a calibration set. In this case, calibration methods are post-processing ~\cite{zadrozny2001obtaining, zadrozny2002transforming,kuleshov2018accurate,platt1999probabilistic,krishnan2020improving}.
Histogram binning~\cite{zadrozny2001obtaining} is a widely used non-parametric calibration method. It groups predictions into confidence-based bins and replaces the model’s confidence with the empirical accuracy of the corresponding bin, producing probabilities that better match true accuracy. Isotonic regression builds on the same idea but learns a flexible monotonic regressor to calibrate confidences, rather than relying on fixed bin boundaries~\cite {kuleshov2018accurate}. Both histogram binning and isotonic regression can be extended to multi-class settings using a one-vs-rest strategy. In contrast, Platt scaling is a parametric calibration method~\cite{platt1999probabilistic, kuppers2020multivariate}. Platt scaling first passes the model's logits to a linear transformation before applying the softmax function. The transformation is learned on a separate calibration set to better match predicted confidence with actual accuracy. A special case is temperature scaling, where the logits are simply multiplied by a scalar~\cite{guo2017calibration,kull2019beyond}. 
However, the methods introduced above are solely based on predicted logits or confidence, without incorporating input features. As a result, they may produce overconfident or underconfident predictions for certain inputs.

Some calibration techniques can be incorporated directly into the training process, without relying on an additional calibration set. A standard classification model is usually trained by minimizing the negative log-likelihood (NLL). However, NLL does not only aim to predict the correct label, as it also pushes the model to assign high confidence to that label for every training example. This pressure to drive the probability of the predicted class toward one can lead to overfitting and, consequently, poor calibration~\cite{mukhoti2020calibrating}.
To address this issue, several training-time methods intentionally soften the confidence that the model assigns during learning. Label smoothing~\cite{muller2019does} does this by slightly reducing the probability of the predicted class and redistributing the reduction uniformly to other classes. In effect, the model is discouraged from becoming overly certain about any single prediction. Similarly, focal loss~\cite{lin2017focal} reduces the influence of examples that the model already predicts with high confidence during training, so that they contribute less to the gradient. This helps prevent the model from becoming excessively confident. Accuracy versus uncertainty calibration (AvUC) loss~\cite{krishnan2020improving} promotes consistency between prediction correctness and uncertainty by rewarding high confidence to correct predictions while penalizing overconfidence on incorrect ones.

\noindent \textbf{Calibration for GNNs.}
Calibration on classification GNNs started to draw research attention recently as well~\cite{teixeira2019graph,wang2021confident,hsu2022makes,liu2022calibration,wang2022gcl,nascimento2024n,zhang2021multi,10445967,trivedi2024accurate,zhang2018link}. 
Traditional calibration methods assume i.i.d. data and thus perform poorly on GNNs, where each node’s behavior depends on its neighbors. These relational dependencies limit the effectiveness of conventional approaches. 
In graph settings, a node’s calibration error is shaped by its neighbors, with well-calibrated confidence distributions often being homophilic~\cite{wang2021confident}, meaning neighboring nodes tend to share similar ground-truth confidences. Calibration error also decreases as the number of same-label neighbors increases~\cite{hsu2022makes}, suggesting that strong local label agreement improves calibration.
Beyond data dependencies, several architectural and training factors also affect GNN calibration. A larger hidden dimensionality per layer generally improves calibration, whereas increasing the number of layers can shift models from underconfidence to overconfidence~\cite{liu2022calibration, wang2022gcl}. Liu et al.~\cite{liu2022calibration} further show that GNNs tend to be under-confident with early stopping but become overconfident when trained for too many epochs. Additionally, models trained on multi-class tasks are typically better calibrated than those trained on binary tasks~\cite{vos2024calibration}.

The calibration methods for GNNs can be post-hoc~\cite{wang2021confident,hsu2022makes,liu2022calibration, yang2022calibrate, yang2024calibrating, nascimento2024n}. 
CaGCN is a topology-aware post-hoc calibration model assuming that neighboring nodes should share similar prediction probability distributions if the model is well-calibrated~\cite{wang2021confident}. To this end, a secondary GCN is built on top of an accuracy-optimized GCN, and it functions like a graph-adapted Platt scaling method to propagate confidence along the network topology. 
Graph Attention Temperature Scaling (GATS)~\cite{hsu2022makes} embeds the attention mechanism for node-wise calibration. For each node, a scaling process adjusts its prediction logit using a global calibration bias and a relative confidence compared against its neighbors. Ratio-binned scaling (RBS)~\cite{liu2022calibration} follows the logic of histogram binning to calibrate outputs of GNNs. RBS first estimates the same-class-neighbor ratio of each node, then groups the nodes into bins according to their ratios, and finally learns a temperature for each bin on a validation set. 
Yang et al.~\cite{yang2024calibrating} observe that calibration performance is more strongly influenced by the topology of the dataset than by the specific choice of GNN architecture. To address this, they propose Data-Centric Graph Calibration (DCGC), which enhances calibration by adjusting the adjacency matrix of the input graph instead of modifying model parameters, thereby mitigating structural biases present in the graph data.
Recent research reveals that GNNs tend to be overconfident in negative link predictions but underconfident in positive ones. 
To address this issue, IN-N-OUT~\cite{nascimento2024n} calibrates link prediction by learning a temperature parameter for each edge via edge perturbation. Each edge embedding is formed by aggregating the embeddings of its incident nodes. During inference, two embeddings are computed for each edge: one with the edge included in message passing and one with it excluded. The discrepancy between these embeddings guides temperature scaling, where a larger difference indicate lower confidence in the edge’s existence.

Conducting calibration during model training is also attractive since it does not require additional data and computation costs.  Graph Calibration Loss (GCL)~\cite{wang2022gcl} enables an end-to-end calibration method for GNNs by minimizing the KL divergence between the ground-truth and predicted confidence distributions. HyperU-GCN~\cite{yang2022calibrate} is proposed to calibrate hyperparameter uncertainty in automated graph learning, where hyperparameter optimization (HPO) is applied to GNNs, by leveraging a bi-level optimization strategy.

\noindent \textbf{Calibration for GFMs.} Calibrating GFMs using external tools can improve reliability. IGDA~\cite{havrilla2025igda} uses LLMs to assign pseudo confidence scores to unlabeled edges, with iteratively refinement based on neighboring contexts. Calibration on foundation models also provides crucial insights for GFMs. Zhu et al.~\cite{zhu2023generalized} posit that pretraining data biases shift finetuned decision boundaries, and calibrate logits by approximating the pretraining label priors and minimizing downstream risks. Hu et al.~\cite{hu2025inference} focus on pretraining-inference mismatch in Visual Language Models (VLMs), and leverage mean and covariance of data distribution to calibrate features for zero-shot and few-shot cross-modal tasks. Zhu et al.~\cite{zhu2023calibration} analyze impacts of training factors on calibrating LLMs. They found that scaling model sizes and training steps benefit calibration, while instructional alignment tuning can impair calibration by disrupting distributions.

\underline{\textit{Summary}}.
Post-hoc calibration methods apply a lightweight adjustment using a few parameters to align a model's predicted confidence with its empirical accuracy, making them highly efficient and model-agnostic compared to computationally intensive Bayesian methods~\cite{wang2021confident,hsu2022makes,liu2022calibration, yang2022calibrate}. However, this parsimony also renders them less flexible, often failing to correct complex GNN miscalibration patterns. Recent work attempts calibration-based loss during training to alleviate the issue~\cite{wang2022gcl}. Yet, the data-centric nature of calibration relies heavily on a representative labeled set, making it particularly vulnerable to performance degradation with OOD data.

\subsection{Conformal Prediction}
\label{sec:conformal_prediction}

\noindent \textbf{Conformal Prediction Overview.}
Given a test input, conformal prediction (CP) generates a prediction set (rather than a single prediction) for each test input using a statistical threshold derived from prediction confidences of calibration data. The prediction set guarantees the inclusion of the true label with a specified probability~\cite{vovk2005algorithmic} or an upper bound of a user-defined prediction loss~\cite{angelopoulos2023conformalriskcontrol}. We focus on CP for classification.


Inductive CP, or split conformal prediction (SCP), is a widely applied version of CP~\cite{papadopoulos2002inductive}. For a trained classifier ${f}$ with label space $\mathcal{Y}=\{1,...,K\}$, 
the calibration set $\mathcal{D}_\text{cal}= \{(X_i,Y_i)\}_{i=1}^n$ is applied to quantify how uncertain the model $f$ would be. Denote $(\Pi_i^{(1)},...,\Pi_i^{(K)})$ the estimated probability for each class from $f(X_i)$. The conformal score for $(X_i, Y_i)\in \mathcal{D}_\text{cal}$ is defined as $s(X_i, Y_i)=1-\Pi_i^{(Y_i)}$, which is widely selected for classification tasks and measures how predictions disagree with the true labels. Let $1-\alpha$ denote the desired confidence level, we define $\tau$ as the $1-\alpha$ quantile of the conformal score set $\mathcal{S}=\{s(X_i,Y_i)\}_{i=1}^n$: 
\begin{equation} \label{eq: S quantile}
\tau=\text{Quantile}\left(1-\alpha,\mathcal{S}\right).
\end{equation}

For a test input $X_{n+1}$, a prediction set $\mathcal{C}(X_{n+1})=\{k|1-\Pi_{n+1}^{(k)}\le \tau, k \in\mathcal{Y}\}$ includes all classes $k\in\mathcal{Y}=\{1,...,K\}$ whose conformal score is less or equal to $\tau$. If the test sample $(X_{n+1}, Y_{n+1})$ is exchangeable with the elements of $\mathcal{D}_\text{cal}$, a coverage guarantee holds that
\begin{equation} \label{eq: prediction set}
\text{Pr}\left(Y_{n+1}\in \mathcal{C}(X_{n+1})\right) \ge 1-\alpha.
\end{equation}
To make the size of $\mathcal{C}(X_{n+1})$ more adaptive to $X_{n+1}$ under the exchangeability assumption, novel conformal score functions are proposed ~\cite{romano2020classification,angelopoulos2020uncertainty,romano2019conformalized,guan2023localized}. However, the exchangeability assumption can be violated by distribution shift between calibration distribution $P({X,Y})$ and test distribution $Q({X,Y})$ such that $(X_{n+1}, Y_{n+1})$ is non-exchangeable with the elements of $\mathcal{D}_\text{cal}$.
In the case of covariate shift ($P(X)\neq Q(X)$), an importance weighting technique is introduced based on the likelihood ratio of $X$ to maintain the coverage guarantee in Eq.~(\ref{eq: prediction set})~\cite{tibshirani2019conformal}. When joint distribution shift ($P({X})\neq Q({X})$, $P({Y|X})\neq Q({Y|X})$) occurs, prediction sets can be adjusted to keep $1-\alpha$ coverage on test data under dynamic distribution shift~\cite{Gibbs2021AdaptiveCI,xu2021conformal} and static distribution shift~\cite{cauchois2024robust, gendler2021adversarially,zou2024coverage}.  It is proved that the coverage gap can be bounded by the total variation distance~\cite{barber2023conformal} and Wasserstein distance~\cite{xu2025wassersteinregularized} between calibration and test conformal scores under the non-exchangeable condition.

\noindent \textbf{Conformal Prediction for GNNs.}
When labeled calibration nodes and test nodes are uniformly distributed throughout the graph, it is reasonable to assume that calibration and test nodes are exchangeable. Under this setting, marginal coverage guarantees remain valid. JuryGCN~\cite{kang2022jurygcn} estimates prediction uncertainty in node classification tasks using jackknife methods. 
In graphs, node neighbors with similar labels (homophily) provide strong predictive cues, leading to higher model confidence, whereas neighbors with differing labels (heterophily) increase prediction uncertainty. Motivated by this observation, Diffused Adaptive Prediction Sets (DAPS)~\cite{zargarbashi2023conformal} better reflect a node's local uncertainty by leveraging neighborhood diffusion in the graph and define the diffused score as:
\begin{equation}
s^\text{diff}(X_i,Y_i) = (1 - \beta) s(X_i,Y_i) + \frac{\beta}{|\mathcal{N}_i|} \sum_{X_j \in \mathcal{N}_i} s(X_j, Y_i),
\end{equation}
Zargarbashi et al.~\cite{zargarbashi2024conformal} further introduce an edge-exchangeable setting to address the sparsity found in realistic calibration graph sets.

However, when calibration and test data correspond to different subgraphs with distinctive topological properties, feature distributions, or label characteristics, the exchangeability assumption typically breaks down.
This issue has been highlighted in ~\cite{zargarbashi2023conformal}, which emphasizes that topological locality and data partitioning within graphs introduce non-exchangeability that vanilla CP methods fail to handle.  To address the non-exchangeability in tabular data, Barber et al.~\cite{barber2023conformal} propose Non-exchangeable Conformal Prediction (NexCP) that assigns deterministic weights to each calibration point. Conformal scores with higher weights effectively occupy a larger “share” when computing the quantile used to form the prediction set. The main challenge is choosing weights that reflect the similarity between calibration samples and the test input. Intuitively, calibration points drawn from distributions closer to that of the test sample should receive higher weights, as they provide more relevant information for estimating uncertainty.
This idea has been extended to graph data by Clarkson et al.~\cite{clarkson2023distribution}, who integrate NexCP into the Adaptive Prediction Sets (APS) framework~\cite{romano2020classification}. 
By graph homophily property,
Neighborhood Adaptive Prediction Sets (NAPS) assign weights to calibration nodes based on their distances to the test node. Nodes that are closer (in terms of hop count or diffusion distance) are considered more representative and are thus given greater influence in determining the prediction set. 

In addition to node classification, CP has been applied to other graph-related tasks, such as link prediction~\cite{marandon2024conformal,zhao2024conformalized,luo2023anomalous}, graph regression~\cite{lunde2023conformal,luo2024conformal} and classification~\cite{wu2024conditionalpredictionrocbands}. These studies largely extend inductive CP to new task settings while still relying on the assumption of exchangeability.

\noindent \textbf{Conformal Prediction for GFMs.} Although no work directly examines CP for GFMs, several studies investigate CP in LLMs that provide insights for GFMs. They are also categorized into "relying on exchangeability" and "beyond exchangeability". In the former, studies leverage intrinsic signals, such as confidence~\cite{gui2024conformal, kumar2023conformal}, self-consistency~\cite{wang2024conu}, or input difficulty~\cite{cherian2024large} to construct prediction sets with finite-sample, distribution-free guarantees. By dynamically calibrating coverage or selecting reliable outputs~\cite{quach2023conformal}, they ensure statistically valid predictions with efficiency. In the latter, approaches focus on reliable UQ under distribution shifts. Techniques such as non-exchangeable sampling provide token-level conformal prediction sets~\cite{ulmer2024non}, and adaptive rejection frameworks handle continual learning and distributional shifts by selectively abstaining or reweighting calibration data~\cite{zhou2025robust}. In summary, intrinsic signal calibration and robust handling of distribution shifts in CP for LLMs provide a technical roadmap for UQ in GFMs.

\underline{\textit{Summary}}. 
Conformal prediction offers a distribution-free framework for providing a prediction set for each test instance using finite calibration samples. The largest challenge in applying it to graph-structured data is the inherent violation of this exchangeability assumption, as the relational structure and node interdependencies create non-i.i.d. distributions between the calibration and test sets, which directly breaks the theoretical coverage guarantee and makes its practical reliability uncertain. Although contemporary research attempts to mitigate this by approximating the extent of distribution shift~\cite{clarkson2023distribution,barber2023conformal}, the fundamental lack of a rigorous coverage assurance for graph data in real-world, non-i.i.d. scenarios remains a significant and unresolved concern, limiting the trustworthiness of the method in high-stakes applications.



\section{Uncertainty Quantification Evaluation}
\label{sec:uq_evaluation}

\subsection{Bayesian Inference Evaluation}

Although Bayesian methods aim to compute or approximate posterior distributions over model parameters or latent variables, their evaluation in graph learning usually does not test posterior correctness directly, since the ground-truth posterior is unavailable in real graph datasets \cite{zhang2019bayesian, stadler2021graph}. Instead, existing work evaluates whether the uncertainty assigned to nodes, edges, or entire graphs is informative about prediction failure or distribution shift \cite{zhao2020uncertainty, hasanzadeh2019semi}.

In misclassification detection, the estimated uncertainty such as predictive entropy or variance is used as a score to distinguish between correct and incorrect predictions. 
An effective inference mechanism should assign significantly higher uncertainty to misclassified nodes or graphs. 
Similarly, for OOD detection, the evaluation measures how well the uncertainty estimates can differentiate between in-distribution and OOD graph data. 
The performance is evaluated using threshold-independent metrics, such as 
AUROC and AUPRC, 
treating the uncertainty estimate as a binary classification score.

Current evaluation protocols for Bayesian graph models mainly assess the \emph{utility} of uncertainty through ranking-based tasks, rather than the \emph{faithfulness} of the inferred posterior.

\subsection{Evaluating Uncertainty due to OOD Data}
DRO, AT, and SSL methods for graphs primarily target generalization under shift and are typically evaluated through \emph{performance under perturbations} rather than through uncertainty-specific criteria. 
DRO is commonly assessed by reporting worst-case performance as the uncertainty-set size increases, e.g., by exposing a budget-performance trade-off between robust node/graph classification accuracy versus the radius $\rho$ of a Wasserstein neighborhood \cite{zhang2021robust,wang2022distributionally,chen2023uncertainty}. 
AT is evaluated by clean accuracy together with adversarial accuracy under a specified attack model and perturbation budget (e.g., feature/edge/topology perturbations), and similarly exhibits an accuracy-robustness trade-off as the attacking budget increases \cite{madry2017towards,gosch2024adversarial}. 
SSL methods are typically evaluated by downstream task performance under distribution shifts induced by alternative augmentation/sampling regimes, measuring stability and robustness across views rather than calibrated predictive uncertainty \cite{you2020graph,qiu2020gcc}.

Robustness-oriented evaluations are important for assessing performance under graph shift, but they do not directly verify whether uncertainty estimates are well calibrated or whether they increase appropriately on structurally novel regions.

\subsection{Calibration}
The evaluation of calibration measures the difference between a model's prediction confidence and the exact probability that the predicted output is correct. Two widely applied metrics are introduced below.

How well a model’s predicted confidences match its actual accuracy can be measured by Expected Calibration Error (ECE). To compute it, predictions are grouped into several confidence bins (for example, $0–10\%$, $10–20\%$, …). For each bin, we calculate the average confidence and the average accuracy, and ECE summarizes the weighted average of the absolute differences between these two quantities. A perfectly calibrated model would have matching accuracy and confidence in every bin, appearing as a diagonal line in a reliability diagram. Variants such as Maximum Calibration Error~\cite{naeini2015obtaining} and classwise ECE~\cite{kull2019beyond} evaluate the largest gap or the gap within each class.
For node-level classification on graphs, nodewise ECE applies the same idea by binning test nodes according to their predicted confidence~\cite{wang2021confident,wang2022gcl,teixeira2019graph,liu2022calibration}. However, this ignores graph structure and correlations between neighboring nodes. To address this, edgewise ECE~\cite{hsu2022graph} evaluates calibration at the level of edges: each edge receives a confidence score based on the confidences of its endpoints, and an edge is counted as correct only if both endpoint predictions are correct. The discrepancy between edge confidences and edge accuracies across bins yields the edgewise ECE. Additional metrics have also been proposed to separately evaluate calibration on edges where the connected nodes agree or disagree in their predictions~\cite{hsu2022graph}.

While ECE depends on how predictions are grouped into bins, the Brier Score provides a continuous and bin-free way to measure calibration~\cite{brier1950verification}. It evaluates the quality of a model’s probabilistic predictions by comparing the full probability distribution assigned to each class with the actual probability that each class is the true label. A lower Brier Score indicates that the model assigns high probability to the correct class and distributes little probability to incorrect ones.

\subsection{Conformal Prediction}

A central property of conformal prediction (CP) is its guarantee of marginal coverage at the nominal level $1-\alpha$. To assess this in practice, we compute the empirical coverage on a held-out test set, which is the fraction of examples whose true label lies within the predicted set, and compare it to the target level. We report the marginal coverage gap, defined as the deviation between empirical coverage and 
$1-\alpha$, to quantify how well the method satisfies its coverage guarantee.


Moreover, it is desired that CP can be adaptive for different inputs (i.e.,$1-\alpha$ coverage holds for different features across the entire input space)~\cite{angelopoulos2020uncertainty}. Worst-Slice Coverage (WSC)~\cite{cauchois2021knowing} is a metric designed for this purpose: it measures the lowest conditional coverage over a collection of input-space slices. Therefore, WSC reveals how well a conformal predictor handles the most underrepresented subspace. To ensure statistical reliability, each slice must be sufficiently large, typically containing at least $10\%$ of the test samples. 

Prediction efficiency is important in practice as well, which refers to how tight the prediction sets are. Smaller prediction sets are more informative for users to identify the true labels. Under guaranteed coverage, evaluating efficiency provides critical insight into the precision of the predictive sets.

\section{Future Directions}
\label{sec:future_direction}
In this section, we analyze existing challenges in uncertainty quantification on graphs, and summarize some future research directions facing these challenges.

\textbf{Bayesian learning} can generate credible regions for model predictions, but these regions are only valid if the model is correctly specified, meaning that the prior, likelihood, and computational capability must be accurately defined during posterior inference \cite{cha2023temperature,masegosa2020learning,wilson2020bayesian}. 
Cha et al. \cite{cha2023temperature} primarily addresses these issues by introducing a temperature parameter into Bayesian GNNs within the conformal prediction framework. 
The combination of Bayesian inference with the conformal prediction framework allows for both credible regions and prediction sets that are more robust, even in the presence of model misspecification.
Integrating Bayesian methods with other UQ techniques could mitigate inaccuracies arising from violations of Bayesian learning assumptions. 
By leveraging the strengths of other UQ methods, such integrations could lead to more robust uncertainty estimates for graph learning.

\textbf{Out-of-distribution} tasks have been well explored by various techniques, including DRO, AT, and SSL. 
As discussed before, the majority of existing research has primarily focused on improving model generalization to OOD samples. 
However, there has been limited exploration into the deeper connections between UQ and OOD.
A promising direction is to leverage OOD methods to better estimate and quantify uncertainties.
Integrating OOD techniques with UQ \cite{de2025deep,linmans2023predictive} could yield valuable insights, potentially leading to more robust and reliable uncertainty estimations. The integration could also help bridge the gap between OOD and UQ.

\textbf{Conformal prediction} methods for graph data concentrate on static graphs to adhere to the exchangeability assumption or fixed distribution shift. 
However, CP for dynamic graphs presents a unique challenge. 
Because dynamic graphs, such as social networks in real-world, can result in dynamic distribution shifts~\cite{davis2024valid}. 
Addressing this challenge could enhance the interpretability of GNNs when dealing with evolving data, like changes in edges over time.

\textbf{Calibration} methods primarily address miscalibration errors by leveraging the homophily property of graphs
\cite{hsu2022makes, wang2021confident}. This approach has proven effective in many scenarios, but it falls short in heterophilous environments where neighboring nodes are more likely to belong to different classes. 
Effectively calibrating GNNs under heterophily is essential for improving their reliability and performance in real-world applications.
However, the calibration of GNNs in these contexts remains underexplored.

\section{Conclusion}
\label{sec:conclusion}

In this survey, we carefully reviewed the uncertainty quantification within graphical models, specifically focusing on Graph Neural Networks and Graph Foundation Models. We examined existing methods for representing and handling uncertainty, including Bayesian representation learning, Gaussian Processes, Outlier Representation, approaches based on Stochastic Differential Equations (SDEs), conformal prediction, calibration, and so on.
The domain of uncertainty in graphical models presents substantial opportunities for further innovation and advancement. Advancing our understanding of uncertainty modeling in graph-based systems will empower machine learning systems to make more reliable decisions in the presence of real-world noise. Such progress will not only improve the accuracy and safety of AI applications but also promote trust and encourage the broader adoption of these technologies in practical settings.


\bibliographystyle{IEEEtran}
\bibliography{main.bib}

\clearpage
\newpage
\begin{appendices}
\begin{figure*}
    \centering
    \begin{forest}
        for tree={
                    forked edges,
                    grow'=0,
                    draw=none,
                    rounded corners,
                    node options={align=center,},
                    text width=2.7cm,
                    anchor=center
                },
                [\small Uncertainty Quantification, fill=gray!10 ,rotate=90, parent
                    [\small Uncertainty Representation Sec.~\ref{sec:method_uncertainty_representation}, for tree={fill=teal!50, child}
                                            [\small Dirichlet--Multinomial Representation Sec.~\ref{sec:dirichlet-motinomial}, fill=teal!30, greatgrandchild
                        [\small {SocNL \cite{yamaguchi2015socnl}, NETCONF \cite{eswaran2017power}, GPN \cite{stadler2021graph}}, fill=teal!10, referenceblock]]
                        [\small Bayesian Representation Learning Sec.~\ref{subsec: bayesian representation learning}, fill=teal!30, greatgrandchild
                        [\small {Graph Variational Autoencoders: VGAE \cite{kipf2016variational}, SIG-VAE \cite{hasanzadeh2019semi}, DGVAE \cite{li2020dirichlet}, VGNAE \cite{ahn2021variational}, Graphite \cite{grover2019graphite}, GraphVAE \cite{simonovsky2018graphvae}, VGRNN \cite{hajiramezanali2019variational}, HVGAE \cite{sun2021hyperbolic}}, fill=teal!10, referenceblock]
                        [\small {Bayesian Graph Neural Networks: Bayesian GCNN \cite{zhang2019bayesian}, BGCN (node copying) \cite{pal2019bayesian}, BGCN (non-parametric) \cite{pal2020non}, BUP \cite{xu2022uncertainty}, GKDE \cite{zhao2020uncertainty}, VGCN \cite{elinas2020variational}, BGNN \cite{munikoti2023general}}, fill=teal!10, referenceblock]]
                        [\small Gaussian Process Sec.~\ref{subsec: gaussain process}, fill=teal!30, greatgrandchild
                        [\small {GGP \cite{ng2018bayesian}, UaGGP \cite{liu2020uncertainty}, LNK \cite{wollschlager2023uncertainty}, GPGC \cite{hu2020infinitely}, WGGP \cite{opolka2022adaptive}, GPG \cite{venkitaraman2020gaussian}, DGPG \cite{li2020stochastic}, FT/WT-GP \cite{opolka2023graph}, GCLGP \cite{opolka2020graph}, GGPN \cite{chen2022multi}}, fill=teal!10, referenceblock]]
                        [\small Outlier Representation Sec.~\ref{sec:outlier_representation}, fill=teal!30, greatgrandchild
                            [\small {Dataset-level uncertainty estimation: 
                            ADNCE \cite{wu2024understanding}, 
                            WDRO \cite{zhang2021robust}, 
                            DRGL \cite{chen2023uncertainty}
                            },
                            fill=teal!10, referenceblock]
                            [\small {Sample-level uncertainty representation: 
                            CAP \cite{xue2021cap},
                            Xu et al. \cite{xu2019topology},
                            DWNS \cite{dai2019adversarial}, 
                            RGIB \cite{wang2023toward} 
                            },
                            fill=teal!10, referenceblock]
                        [\small {Manually designed uncertainty sets: 
                            GraphCL \cite{you2020graph}, 
                            Hu et al. \cite{hu2019strategies}, 
                            GRACE \cite{zhu2020deep}
                            },
                            fill=teal!10, referenceblock]
                        ]
                        [\small Stochastic Differential Equations Sec.~\ref{sec:SDEs_representation}, fill=teal!30, greatgrandchild
                        [{\small LGNSDE \cite{bergna2025uncertainty}, GNSD \cite{lin2024graph}, DVGNN \cite{liang2024dynamic}, Graphgdp \cite{huang2022graphgdp}, AGGDN \cite{xing2023aggdn}},fill=teal!10, referenceblock]]
                    ]
                    [\small Uncertainty Handling Sec.~\ref{sec:method_uncertainty_handling}, for tree={fill=orange!45,child}
                        [\small Bayesian Inference Techniques Sec.~\ref{subsec: bayesian_inference_techniques}, fill=orange!30, greatgrandchild
                        [\small {Exact: SocNL \cite{yamaguchi2015socnl}, NETCONF \cite{eswaran2017power}, GPN \cite{stadler2021graph}}, fill=orange!10, referenceblock]
                        [\small {Approximate: VGAE \cite{kipf2016variational}, SIG-VAE \cite{hasanzadeh2019semi}, DGVAE \cite{li2020dirichlet}, VGNAE \cite{ahn2021variational}, Graphite \cite{grover2019graphite}, GraphVAE \cite{simonovsky2018graphvae}, VGRNN \cite{hajiramezanali2019variational}, HVGAE \cite{sun2021hyperbolic},Bayesian GCNN \cite{zhang2019bayesian}, BGCN (node copying) \cite{pal2019bayesian}, BGCN (non-parametric) \cite{pal2020non}, GKDE \cite{zhao2020uncertainty}, VGCN \cite{elinas2020variational}, BGNN \cite{munikoti2023general}, GGP \cite{ng2018bayesian}, UaGGP \cite{liu2020uncertainty}, LNK \cite{wollschlager2023uncertainty}, WGGP \cite{opolka2022adaptive}, DGPG \cite{li2020stochastic}, FT/WT-GP \cite{opolka2023graph}, GCLGP \cite{opolka2020graph}, GGPN \cite{chen2022multi}, LGNSDE \cite{bergna2025uncertainty}, DVGNN \cite{liang2024dynamic}, }, fill=orange!10, referenceblock]
                        ]
                        [\small Out of Distribution Techniques Sec.~\ref{sec:ood_handling}, fill=orange!30, greatgrandchild
                            [\small {Distributionally Robust Optimization:
                                ADNCE \cite{wu2024understanding}, 
                                DR-GNN \cite{wang2024distributionally}, 
                                WDRO \cite{zhang2021robust}, 
                                DRGL \cite{chen2023uncertainty},
                                Sadeghi, et al. \cite{sadeghi2021distributionally}
                                },
                                fill=orange!10, referenceblock]
                            [\small {Adversarial Training:
                                WT-AWP \cite{wu2023adversarial}, CAP \cite{xue2021cap}
                                PR-BCD \cite{geisler2021robustness},
                                GraphDefense \cite{wang2019graphdefense},
                                DWNS \cite{dai2019adversarial}, 
                                BVAT \cite{deng2023batch}, 
                                GraphAT \cite{feng2019graph}, 
                                RGIB \cite{wang2023toward} 
                                },
                                fill=orange!10, referenceblock]
                            [\small {Self-Supervised Learning:
                                GraphCL \cite{you2020graph}, 
                                GCC \cite{qiu2020gcc},
                                SS-GCN \cite{you2020does}, 
                                Hu et al. \cite{hu2019strategies}, 
                                GRACE \cite{zhu2020deep}
                                },
                                fill=orange!10, referenceblock]
                            ]
                        [\small Calibration Sec.~\ref{sec:calibration}, fill=orange!30, greatgrandchild
                        [{\small CaGCN~\cite{wang2021confident}, GCL~\cite{wang2022gcl}, MCCNIFTY~\cite{zhang2021multi}, IN-N-OUT~\cite{nascimento2024n}, E-$\Delta$UQ~\cite{10445967}, G-$\Delta$UQ~\cite{trivedi2024accurate}},fill=orange!10, referenceblock]]
                        [\small Conformal Prediction Sec.~\ref{sec:conformal_prediction}, fill=orange!30, greatgrandchild
                        [{\small JuryGCN~\cite{kang2022jurygcn}, DAPS~\cite{zargarbashi2023conformal}, NAPS~\cite{clarkson2023distribution}},fill=orange!10, referenceblock] ]
                        ]
                    ]
    \end{forest}
    \caption{The taxonomy of uncertainty quantification methods in graph learning is organized into two main branches: Uncertainty Representation and Uncertainty Handling. Uncertainty Representation addresses how uncertainty is captured within data and models, including methods such as Bayesian Representation Learning, Gaussian Processes, and Outlier Representation. Uncertainty Handling focuses on methods for evaluating and mitigating uncertainty, encompassing Direct Inference, Out-of-Distribution Techniques, Calibration, and Conformal Prediction. Each leaf node highlights representative works and associated references for each method.}
    \label{fig: Tree diagram}
\end{figure*}

\section{Resources}
\label{sec: resources}

For convenient access, Table \ref{tab:open_source_code} and Table \ref{tab:tutorials_uq} summarize open-source code repositories from some cited papers on related topics and tutorials from prominent machine learning conferences, respectively.
Fig. \ref{fig: Tree diagram} presents an overview of key methods mentioned in the survey.


\begin{table*}[h]
\centering
\caption{Repositories of Open-source Codes}
\def\arraystretch{1.2}
\begin{tabular}{|c|c|c|l|}
\hline
Method & Model & Algorithm & Code Link\\ 
\hline
\multirow{14}{*}{\makecell{Bayesian\\Method}} & \multirow{2}{*}{\makecell{General\\Model}} & VAEs \cite{kingma2013auto} & \url{https://github.com/AntixK/PyTorch-VAE}\\ 
\cline{3-4}
& & BNNs \cite{lampinen2001bayesian,titterington2004bayesian} & \url{https://github.com/Harry24k/bayesian-neural-network-pytorch} \\ 
\cline{2-4}
& \multirow{18}{*}{\makecell{Graphic\\Model}}& NETCONF \cite{eswaran2017power} & \url{https://dhivyaeswaran.github.io/code/netconf.zip} \\
\cline{3-4}
&& GPN \cite{stadler2021graph} & \url{https://www.daml.in.tum.de/graph-postnet} \\
\cline{3-4}
&& VGAE \cite{kipf2016variational} & \url{https://github.com/tkipf/gae} \\
\cline{3-4}
&& DGVAE \cite{li2020dirichlet} & \url{https://github.com/xiyou3368/DGVAE} \\
\cline{3-4}
&& SIG-VAE \cite{hasanzadeh2019semi} & \url{https://github.com/sigvae/SIGraphVAE} \\
\cline{3-4}
&& VGNAE \cite{ahn2021variational} & \url{https://github.com/SeongJinAhn/VGNAE} \\
\cline{3-4}
&& Graphite \cite{grover2019graphite} & \url{https://github.com/ermongroup/graphite} \\
\cline{3-4}
&& VGRNN \cite{hajiramezanali2019variational} & \url{https://github.com/VGraphRNN/VGRNN} \\
\cline{3-4}
&& Bayesian GCNN \cite{zhang2019bayesian} & \url{https://github.com/huawei-noah/BGCN} \\
\cline{3-4}
&& \makecell{Bayesian GCNN \\ (node copying) \cite{pal2019bayesian}}  & \url{https://github.com/floregol/BGCN_copying} \\
\cline{3-4}
&& S-BGCN-T-K \cite{zhao2020uncertainty} & \url{https://github.com/zxj32/uncertainty-GNN} \\
\cline{3-4}
&& VGCN \cite{elinas2020variational} & \url{https://github.com/ebonilla/VGCN} \\
\cline{3-4}
&& GGP \cite{ng2018bayesian} & \url{https://github.com/yincheng/GGP} \\
\cline{3-4}
&& GPG \cite{venkitaraman2020gaussian} & \url{https://www.kth.se/ise/research/reproducibleresearch1.433797} \\
\cline{3-4}
&& DGPG \cite{li2020stochastic} & \url{https://github.com/naiqili/DGPG} \\
\cline{3-4}
&& FT-GP \& WT-GP \cite{opolka2023graph} & \url{https://github.com/FelixOpolka/Graph-Classification-Gaussian-Processes-via-Spectral-Features} \\
\cline{3-4}
&& LNK \cite{wollschlager2023uncertainty} & \url{https://github.com/wollschl/uncertainty_for_molecules} \\
\cline{3-4}
&& GGPN \cite{chen2022multi} & \url{https://github.com/sysu-gzchen/GGPN} \\
\hline
\multirow{13}{*}{\makecell{Conformal\\Prediction}} & \multirow{9}{*}{\makecell{General\\Model}}
& CPCS \cite{tibshirani2019conformal} & \url{https://github.com/ryantibs/conformal/tree/master/tibshirani2019} \\ 
\cline{3-4} 
&& CQR~\cite{romano2019conformalized} & \url{https://github.com/yromano/cqr} \\ 
\cline{3-4} 
&& APS~\cite{romano2020classification} & \url{https://github.com/msesia/arc} \\ 
\cline{3-4} 
&& conformalbayes \cite{Barber2019PredictiveIW} & \url{https://github.com/CoryMcCartan/conformalbayes} \\ 
\cline{3-4}
&& EnbPI \cite{xu2021conformal} & \url{https://github.com/hamrel-cxu/EnbPI} \\
\cline{3-4}
&& \makecell{CP with Missing\\Values \cite{zaffran2023conformal}}  & \url{https://github.com/mzaffran/ConformalPredictionMissingValues} \\
\cline{3-4}
&& LCP \cite{guan2023localized} & \url{https://github.com/LeyingGuan/LCP} \\
\cline{3-4}
&& stabCP \cite{ndiaye2022stable} & \url{https://github.com/EugeneNdiaye/stable_conformal_prediction} \\
\cline{3-4}
&& OQR \cite{feldman2021improving} & \url{https://github.com/Shai128/oqr} \\
\cline{2-4}
& \multirow{4}{*}{\makecell{Graphic\\Model}}& NAPS \cite{clarkson2023distribution} & \url{https://github.com/jase-clarkson/graph\_cp} \\ 
\cline{3-4} 
&& DAPS \cite{zargarbashi2023conformal} & \url{https://github.com/soroushzargar/DAPS} \\ 
\cline{3-4} 
&& CF-GNN \cite{huang2023uncertainty} & \url{https://github.com/snap-stanford/conformalized-gnn} \\ 
\cline{3-4} 
&& JuryGCN \cite{kang2022jurygcn} & \url{https://github.com/BlueWhaleZhou/JuryGCN\_UQ} \\ 
\hline
\multirow{10}{*}{Calibration}&\multirow{5}{*}{\makecell{General\\Model}}
& Dirichlet Calibration \cite{kull2019beyond} & \url{https://github.com/dirichletcal/dirichletcal.github.io} \\
\cline{3-4}
&& Calibrated Regression \cite{kuleshov2018accurate} & \url{https://github.com/AnthonyRentsch/calibrated_regression} \\
\cline{3-4}
&& Focal Calibration \cite{mukhoti2020calibrating} & \url{https://github.com/torrvision/focal_calibration} \\
\cline{3-4}
&& Label Smoothing \cite{muller2019does} & \url{https://github.com/seominseok0429/label-smoothing-visualization-pytorch} \\
\cline{3-4}
&& Spline Calibration \cite{gupta2021calibration} & \url{https://github.com/kartikgupta-at-anu/spline-calibration} \\
\cline{2-4}
&\multirow{5}{*}{\makecell{Graphic\\Model}}
& G-$\Delta$UQ \cite{trivedi2024accurate} & \url{https://github.com/pujacomputes/gduq} \\
\cline{3-4} 
&& CaGCN \cite{wang2021confident} & \url{https://github.com/BUPT-GAMMA/CaGCN} \\ 
\cline{3-4} 
&& GATS \cite{hsu2022makes} & \url{https://github.com/hans66hsu/GATS} \\ 
\cline{3-4} 
&& RBS \cite{liu2022calibration} & \url{https://github.com/liu-yushan/calGNN} \\ 
\cline{3-4} 
&& HyperU-GCN \cite{yang2022calibrate} & \url{https://github.com/xyang2316/HyperU-GCN} \\ 
\hline
\end{tabular}
\label{tab:open_source_code}
\end{table*}

\begin{table*}
\centering
\caption{Tutorials and Workshops on Uncertainty Quantification and Graphic Models}
\def\arraystretch{1.2}
\begin{tabular}{|p{11.5cm}|c|c|c|c|c|c|c|}
\hline
Tutorial/Workshop & $G$ & Bayes & CP & Cal & OOD & Venues & Year \\ 
\hline
\href{https://gnn.seas.upenn.edu/aaai-2025/}{Graph Neural Networks: Architectures, Fundamental Properties and Applications}&\checkmark & & & & & AAAI & 2025
\\
\hline
\href{https://sites.google.com/view/kdd25-gfm-tutorial}{
Graph Foundation Models: Challenges, Methods, and Open Questions}
&\checkmark & & & & & KDD & 2025
\\
\hline
\href{https://iclr.cc/virtual/2025/workshop/23965}{
Quantify Uncertainty and Hallucination in Foundation Models: The Next Frontier in Reliable AI}
& & \checkmark & \checkmark & \checkmark & \checkmark & ICLR & 2025
\\
\hline
Calibration and Bias in Algorithms, Data, and Models
&  &  &  & \checkmark &  & ICML & 2025 \\ 
\hline
\href{https://icml.cc/virtual/2024/tutorial/35233}{Graph Learning: Principles, Challenges, and Open Directions} & \checkmark &  &  &  &  & ICML & 2024 \\ 
\hline
\href{https://alregib.ece.gatech.edu/aaai-2024-tutorial/}{Formalizing Robustness in Neural Networks: Explainability, Uncertainty, and Intervenability} &  &  &  &  & \checkmark & AAAI & 2024 \\ \hline
\href{https://icml.cc/virtual/2024/tutorial/35231}{Distribution-Free Predictive Uncertainty Quantification: Strengths and Limits of Conformal Prediction} &  &  & \checkmark &  &  & ICML & 2024 \\ \hline
\href{https://graph-neural-networks.github.io/tutorial\_kdd23.html}{Graph Neural Networks: Foundation, Frontiers and Applications} & \checkmark &  &  &  &  & KDD & 2023 \\ \hline
\href{https://dl.acm.org/doi/10.1145/3580305.3599565}{Large-Scale Graph Neural Networks: The Past and New Frontiers} & \checkmark &  &  &  &  & KDD & 2023 \\ \hline
\href{https://dl.acm.org/doi/10.1145/3580305.3599562}{Hyperbolic Graph Neural Networks: A Tutorial on Methods and Applications} & \checkmark &  &  &  &  & KDD & 2023 \\ \hline
\href{https://dl.acm.org/doi/10.1145/3580305.3599555}{Fairness in Graph Machine Learning: Recent Advances and Future Prospectives} & \checkmark &  &  &  &  & KDD & 2023 \\ \hline
\href{https://graph-neural-networks.github.io/tutorial\_aaai23.html}{Graph Neural Networks: Foundation, Frontiers and Applications} & \checkmark &  &  &  &  & AAAI & 2023 \\ \hline
\href{https://mtcazzolato.github.io/tutorial-icdm23/}{Temporal Graph Mining for Fraud Detection} & \checkmark &  &  &  &  & ICDM & 2023 \\ \hline
\href{https://bayesopt-tutorial.github.io/}{Recent Advances in Bayesian Optimization} &  & \checkmark &  &  &  & AAAI & 2023 \\ \hline
\href{https://lingkai-kong.com/kdd23\_tutorial/}{Uncertainty Quantification in Deep Learning} &  & \checkmark & \checkmark & \checkmark &  & KDD & 2023 \\ \hline
\href{https://neurips.cc/virtual/2023/tutorial/73953}{Modeling and Exploiting Data Heterogeneity under Distribution Shifts} &  &  &  &  & \checkmark & NeurIPS & 2023 \\ \hline
\href{https://ai.tencent.com/ailab/ml/twgl/}{Trustworthy Graph Learning: Reliability, Explainability, and Privacy Protection} & \checkmark &  &  &  &  & KDD & 2022 \\ \hline
\href{https://dl.acm.org/doi/abs/10.1145/3534678.3542599}{Algorithmic Fairness on Graphs: Methods and Trends} & \checkmark &  &  &  &  & KDD & 2022 \\ \hline
\href{https://quanmingyao.github.io/AutoML.github.io/aaai22-tutorial.html}{Automated Learning from Graph-Structured Data} & \checkmark &  &  &  &  & AAAI & 2022 \\ \hline
\href{https://yushundong.github.io/ICDM\_2022\_tutorial.html}{Fairness in Graph Mining: Metrics, Algorithms, and Applications} & \checkmark &  &  &  &  & ICDM & 2022 \\ \hline
\href{https://neurips.cc/virtual/2022/tutorial/55806}{Advances in Bayesian Optimization} &  & \checkmark &  &  &  & NeurIPS & 2022 \\ \hline
\href{https://bayesopt-tutorial.github.io/syllabus/combined.pdf}{Bayesian Optimization: From Foundations to Advanced Topics} &  & \checkmark &  &  &  & AAAI & 2022 \\ \hline
\href{https://icml.cc/virtual/2022/tutorial/18437}{Sampling as First-Order Optimization over a space of probability measures } &  &  &  &  &  & ICML & 2022 \\ \hline
\href{https://kdd2021graph.github.io/}{Graph Representation Learning: Foundations, Methods, Applications and Systems} & \checkmark &  &  &  &  & KDD & 2021 \\ \hline
\href{https://dl.acm.org/doi/10.1145/3447548.3470794}{Toward Explainable Deep Anomaly Detection} &  &  &  &  & \checkmark & KDD & 2021 \\ \hline
\href{https://neurips.cc/virtual/2021/tutorial/21890}{The Art of Gaussian Processes: Classical and Contemporary} &  & \checkmark &  &  &  & NeurIPS & 2021 \\ \hline
\href{https://iclr.cc/virtual/2021/3912}{On Calibration and Out-of-Domain Generalization} &  &  &  & \checkmark & \checkmark & ICLR & 2021 \\ \hline
\href{https://ai.tencent.com/ailab/ml/KDD-Deep-Graph-Learning.html}{Deep Graph Learning: Foundations, Advances and Applications} & \checkmark &  &  &  &  & KDD & 2020 \\ \hline
\href{http://www.calvinzang.com/DDLG\_AAAI\_2020.html}{Differential Deep Learning on Graphs and its Applications} & \checkmark &  &  &  &  & AAAI & 2020 \\ \hline
\href{https://icml.cc/virtual/2020/tutorial/5750}{Bayesian Deep Learning and a Probabilistic Perspective of Model Construction} &  & \checkmark &  &  &  & ICML & 2020 \\ \hline
\href{https://raghavchalapathy.github.io/KDD-Tutorials-2020-Deep-Robust-Anomaly-Detection/}{Robust Deep Learning Methods for Anomaly Detection.} &  &  &  &  & \checkmark & KDD & 2020 \\ \hline
\href{https://github.com/nplan-io/kdd2020-calibration}{How to calibrate your neural network classifier: Getting true probabilities from a classification model} &  &  &  & \checkmark &  & KDD & 2020 \\ \hline
\href{https://neurips.cc/virtual/2020/tutorial/16649}{Practical Uncertainty Estimation and Out-of-Distribution Robustness in Deep Learning} &  &  &  &  & \checkmark & NeurIPS & 2020 \\ \hline
\href{https://www.cs.mcgill.ca/$\sim$wlh/grl\_book/}{Graph Representation Learning} & \checkmark &  &  &  &  & AAAI & 2019 \\ \hline
\href{https://iclr.cc/virtual/2019/workshop/631}{Representation Learning on Graphs and Manifolds} & \checkmark &  &  &  &  & ICLR & 2019 \\ \hline
\href{https://icml.cc/virtual/2019/tutorial/4338}{A Primer on PAC-Bayesian Learning} &  & \checkmark &  &  &  & ICML & 2019 \\ \hline
\href{https://neurips.cc/virtual/2019/tutorial/13205}{Deep Learning with Bayesian Principles} &  & \checkmark &  &  &  & NeurIPS & 2019 \\ \hline
\href{https://dl.acm.org/doi/10.1145/3292500.3332267}{Deep Bayesian Mining, Learning and Understanding} &  & \checkmark &  &  &  & KDD & 2019 \\ \hline
\href{http://chien.cm.nctu.edu.tw/home/aaai-tutorial/}{Deep Bayesian and Sequential Learning} &  & \checkmark &  &  &  & AAAI & 2019 \\ \hline
\href{https://icml.cc/virtual/2018/tutorial/1858}{Variational Bayes and Beyond: Bayesian Inference for Big Data} &  & \checkmark &  &  &  & ICML & 2018 \\ \hline
\href{https://neurips.cc/virtual/2018/tutorial/10984}{Scalable Bayesian Inference} &  & \checkmark &  &  &  & NeurIPS & 2018 \\ \hline
\href{https://iclr.cc/virtual/2018/workshop/474}{Bayesian Incremental Learning for Deep Neural Networks} &  & \checkmark &  &  &  & ICLR & 2018 \\ \hline
\href{https://www.microsoft.com/en-us/research/video/bayesian-approaches-for-black-box-optimization/}{Bayesian Approaches for Blackbox Optimization} &  & \checkmark &  &  &  & UAI & 2018 \\ \hline
\href{https://neurips.cc/virtual/2017/tutorial/8731}{Deep Probabilistic Modelling with Gaussian Processes} &  & \checkmark &  &  &  & NeurIPS & 2017 \\ \hline
\href{https://www.researchgate.net/publication/321049891\_Non-IID\_Learning}{Non-IID Learning} &  &  &  &  & \checkmark & KDD & 2017 \\ \hline
\href{https://icml.cc/2015/index.html\%3Fp=97.html}{Bayesian Time Series Modeling: Structured Representations for Scalability} &  & \checkmark &  &  &  & ICML & 2015 \\ \hline
\href{https://www.auai.org/uai2015/proceedings/slides/UAI2015\_LearningBN\_PartI.pdf}{Optimal Algorithms for Learning Bayesian Network Structures} &  & \checkmark &  &  &  & UAI & 2015 \\ \hline
\href{https://www.auai.org/uai2015/proceedings/slides/UAI2015\_Comp\_LN.pdf}{Computational Complexity of Bayesian Networks} &  & \checkmark &  &  &  & UAI & 2015 \\ \hline
\href{https://icml.cc/2014/index/article/17.htm}{Bayesian Posterior Inference in the Big Data Arena: An introduction to probabilistic programming} &  & \checkmark &  &  &  & ICML & 2014 \\ \hline
\href{https://neurips.cc/virtual/2013/tutorial/3686}{Approximate Bayesian Computation (ABC)} &  & \checkmark &  &  &  & NeurIPS & 2013 \\ \hline
\href{https://neurips.cc/virtual/2011/tutorial/2509}{Linear Programming Relaxations for Graphical Models} & \checkmark &  &  &  &  & NeurIPS & 2011 \\ \hline
\href{https://neurips.cc/virtual/2011/tutorial/2506}{Modern Bayesian Nonparametrics} &  & \checkmark &  &  &  & NeurIPS & 2011 \\ \hline
\end{tabular}
\label{tab:tutorials_uq}
\end{table*}

\end{appendices}

\end{document}